\def\includeapmsupplement{1}

\documentclass[10pt,twocolumn,letterpaper]{article}
\usepackage[T1]{fontenc}

\usepackage[pagenumbers]{cvpr} 

\usepackage{placeins}
\newcommand{\apm}{APM\xspace}

\usepackage{colortbl}
\colorlet{rankfirst}{red!30}
\colorlet{ranksecond}{orange!35}
\colorlet{rankthird}{yellow!45}
\newcommand{\best}[1]{\cellcolor{rankfirst}\textbf{#1}}
\newcommand{\secondbest}[1]{\cellcolor{ranksecond}#1}
\newcommand{\thirdbest}[1]{\cellcolor{rankthird}#1}
\newcommand{\rankkey}[2]{{\setlength{\fboxsep}{1pt}\colorbox{#1}{\strut #2}}}  

\makeatletter
\renewcommand\paragraph{\@startsection{paragraph}{4}{\z@}%
  {0.5ex \@plus0.3ex \@minus0.2ex}{-1em}{\normalfont\normalsize\bfseries}}
\makeatother

\ifdefined\includeapmsupplement\else  \usepackage{xr}  \fi

\definecolor{cvprblue}{rgb}{0.21,0.49,0.74}
\usepackage[pagebackref,breaklinks,colorlinks,allcolors=cvprblue]{hyperref}

\def\confName{CVPR}
\def\confYear{2027}

\title{Adversarial Procedural Modeling: Learning Programs for Building Interiors}

\usepackage{orcidlink}
\newcommand{\apmauthors}{%
Fedor Rodionov$^{1}$\,\orcidlink{0009-0009-4306-0154}\quad
Christian Stippel$^{2}$\,\orcidlink{0000-0003-0482-902X}\quad
Michael Birsak$^{1}$\,\orcidlink{0000-0001-6375-8124}\quad
Florian Rist$^{1,2}$\,\orcidlink{0000-0002-1559-0993}\\
Ivan Viola$^{1}$\,\orcidlink{0000-0003-4248-6574}\quad
John Femiani$^{3}$\,\orcidlink{0000-0002-0924-6686}\quad
Peter Wonka$^{1}$\,\orcidlink{0000-0003-0627-9746}\\[0.6em]
{\normalsize $^{1}$King Abdullah University of Science and Technology (kaust.edu.sa)}\\
{\normalsize $^{2}$Technische Universität Wien (tuwien.ac.at)\quad
$^{3}$Miami University of Ohio (miamioh.edu)}%
}

\author{\apmauthors}

\begin{document}
\twocolumn[{%
\renewcommand\twocolumn[1][]{#1}%
\maketitle
\begin{center}
  \captionsetup{type=figure}
  \includegraphics[width=\textwidth]{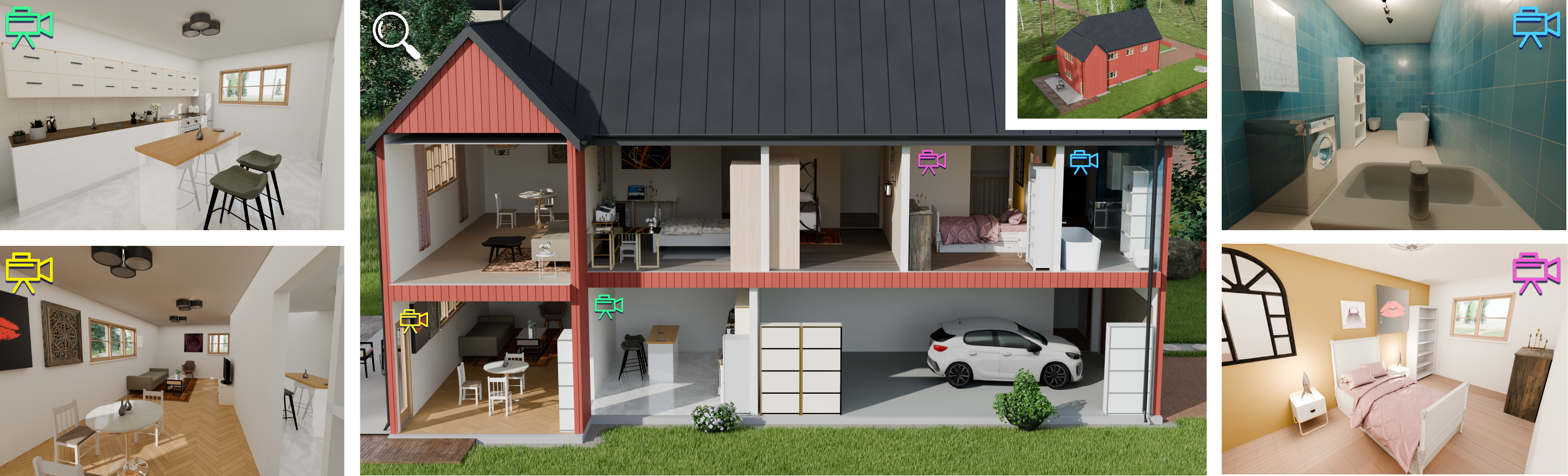}
  \captionof{figure}{\textbf{A home generated by \apm.} The learned procedural program builds the complete two-story home, from floor plan to furniture and decoration: a cut-away view of both stories (inset: exterior) and four interior views, with icons marking where each camera stands. APM does not use neural networks at inference time and can generate a large dataset of buildings very quickly in high quality.}
  \label{fig:teaser_top}
\end{center}%
}]
\begin{abstract}
We propose \emph{Adversarial Procedural Modeling} (\apm)\footnotemark: a framework in which a Generator Agent uses a language-model backend to write and revise a reusable probabilistic program rather than construct each 3D scene through individual placement instructions. The generator runtime executes the probabilistic program via an interpreter and can generate a wide variety of scenes through random sampling.
A separate Discriminator Agent writes code that checks generated batches of 3D scenes against constraints and reference statistics, and provenance records trace each failure back to the source line that produced it.
The language-model cost is paid once, during development.
We demonstrate a realization of this principle for generating building interiors. 
Generation proceeds from a task specification through an architectural program (a semantic graph), architectural layout, furniture placement, and asset placement.
Our results show large improvements in layout quality and generation speed.
Compared to Infinigen Indoors, 70\% of users prefer our results in a blind pairwise comparison, while our generation speed is 450$\times$ faster.
\end{abstract}
    
\begin{figure*}[t]
\centering
\includegraphics[width=\textwidth]{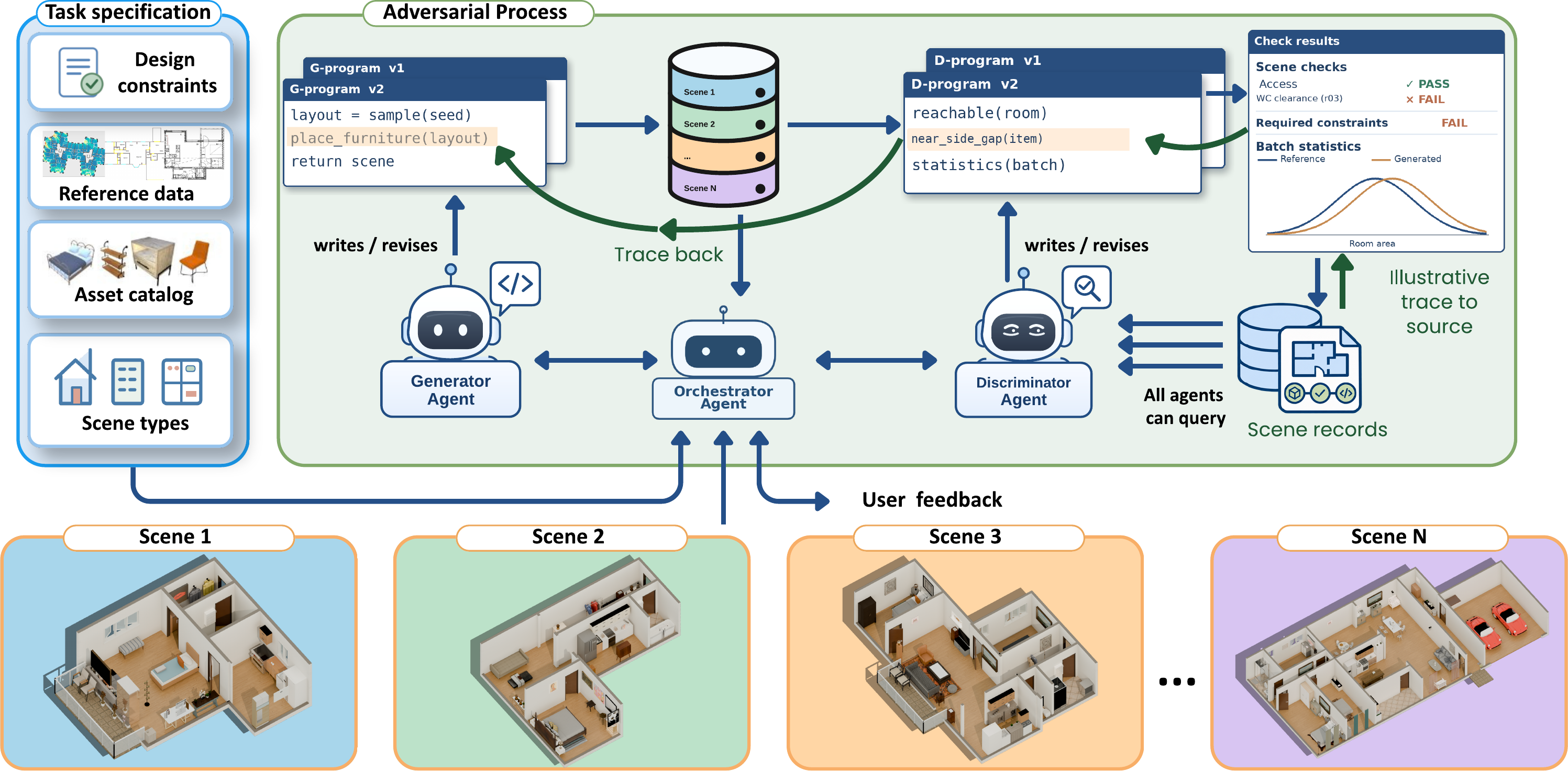}
\caption{\textbf{Adversarial procedural modeling.} The Generator Agent writes and revises a probabilistic G-program, which produces batches of scenes from different random seeds. The Discriminator Agent develops a D-program that checks individual scenes and compares batch statistics with reference data. An Orchestrator Agent directs revisions using the recorded results, selected visual reviews, and optional user feedback. Scene records connect generated elements and check results to their construction histories. The green arrows illustrate how a failed clearance check can lead from the affected element to the associated generator code. Bottom: rendered examples, which the orchestrator may inspect using a vision-capable model to identify missing or incorrect checks.}
\label{fig:teaser}
\end{figure*}

\section{Introduction}
\label{sec:intro}
\footnotetext{Project page: \url{https://OldDeLorean.github.io/APM/}}
Generating building interior layouts is an important instance of generative layout synthesis.
The output should be a 3D scene that includes the room layout, the placement of doors and windows, furniture, and decorative assets (See \cref{fig:teaser_top}).
Such a 3D scene is useful for virtual worlds, robotics simulations, or generating other deep learning training data.
Our goal is to generate varied, plausible apartments and multi-story homes that meet user-specified constraints and preserve the statistics of curated input datasets.

One set of techniques builds on generative modeling using auto-regressive models~\cite{para2021generative, paschalidou2021atiss} or diffusion~\cite{shabani2023housediffusion, tang2024diffuscene}. These techniques are typically limited by the lack of available data.
There aren't any publicly available datasets of complete furnished buildings or apartments with only minor exceptions~\cite{khanna2024habitat, fu20213dfront}\footnote{While 3D-FRONT provides some furnishing, critical rooms such as kitchens and bathrooms are unpopulated.}. The data quality of existing datasets is also questionable, and the data often lacks detailed semantic labels and an absolute scale (e.g., meters).

Therefore, procedural generation offers a promising alternative for this task. A procedural model expressed as a probabilistic program can be inspected, edited, and sampled repeatedly without demanding hardware.
However, current systems such as ProcTHOR~\cite{deitke2022procthor} rely on numerous manually implemented heuristics. This leads to fast generation times but also to substantial limitations in quality and variability.
To improve upon the quality issue, current SOTA relies on hybrid systems and combines procedural modeling with optimization, as Infinigen Indoors~\cite{raistrick2024infinigen}.
The result can improve quality, but modeling times of over 1 h per 3D scene lose a core advantage of procedural modeling. Finally, programmatic rules can be replaced by agent-driven search through layout modifications with manually implemented constraint checks, as in SceneSmith~\cite{pfaff2026scenesmith}. This work avoids reliance on training data but also introduces a new limitation: a high per-scene modeling cost of up to 50 USD. Thus, the fundamental issues of layout quality, scale, and variability still leave ample room for improvement.
To overcome the limitations of previous work, we propose the first modeling framework that 1) generates high-quality building layouts in terms of individual layout quality 2) matches the statistics and diversity of datasets and 3) is cheap in terms of inference time and monetary cost.

We propose \emph{Adversarial Procedural Modeling} (\apm), a new approach for procedural modeling, illustrated in \cref{fig:teaser}.
The \emph{Generator Agent} writes and revises a reusable probabilistic program (\emph{procedural model}) using an LLM backend. 
The probabilistic program produces different building layouts for different random seeds and can generate a complete dataset with large variability.
The \emph{Discriminator Agent} writes and revises a D-program that checks constraints and compares statistics from generated batches of 3D scenes and real datasets.
The agents can use scene records to trace assessment results to the code associated with the affected elements.
Conceptually, the probabilistic program plays the role of model parameters, while the D-program produces an evaluation signal and scene records link individual results to relevant code locations.
The Generator and Discriminator Agents act as optimizers, using this feedback to revise the procedural model.

For example, the generator may satisfy room counts while producing unusual room areas.
The Discriminator Agent can expose this repeated pattern even if no individual sample violates a geometric rule.
The Generator Agent then revises the code part that creates the pattern, rather than repairing only the individual examples.

A unique aspect of our work is that we use two competing agents, each of which writes code.
The generator writes procedural modeling code and the discriminator writes code to extract and assess statistics and constraints.
This is in contrast to many existing approaches that use a loop of design and critic agents operating on a layout directly.
The second unique aspect of our work is that the procedural model is, itself, an output of our system.
We produce programs of substantial size (about 15K lines of Python and 188K lines of JSON) that can generate a large dataset of 3D scenes. 
In summary, we make the following contributions:
\begin{itemize}
\item We propose \emph{adversarial procedural modeling}, a new modeling paradigm for procedural code generation. In contrast to existing solutions that use agents to write one procedural program per 3D scene, we use agents to generate a single complex procedural program that is able to generate millions of 3D scenes.
\item We propose the first method that \emph{quickly} synthesizes \emph{complete} building layouts in \emph{high quality} and \emph{variability}. In blinded pairwise comparisons, human participants prefer \apm over every baseline in 70--93\% of the pairs, and two vision-language model judges in 67--100\% (\cref{tab:userstudy}). Sampling a complete furnished home takes 0.39\,min on one CPU core without API cost, over 450$\times$ faster than Infinigen Indoors (178.4\,min per home); SceneSmith needs 206\,min for a single room, as reported by its authors.
\end{itemize}

\section{Related work}
\label{sec:related}

\begin{figure*}[tb!]
  \centering
  \includegraphics[width=\textwidth]{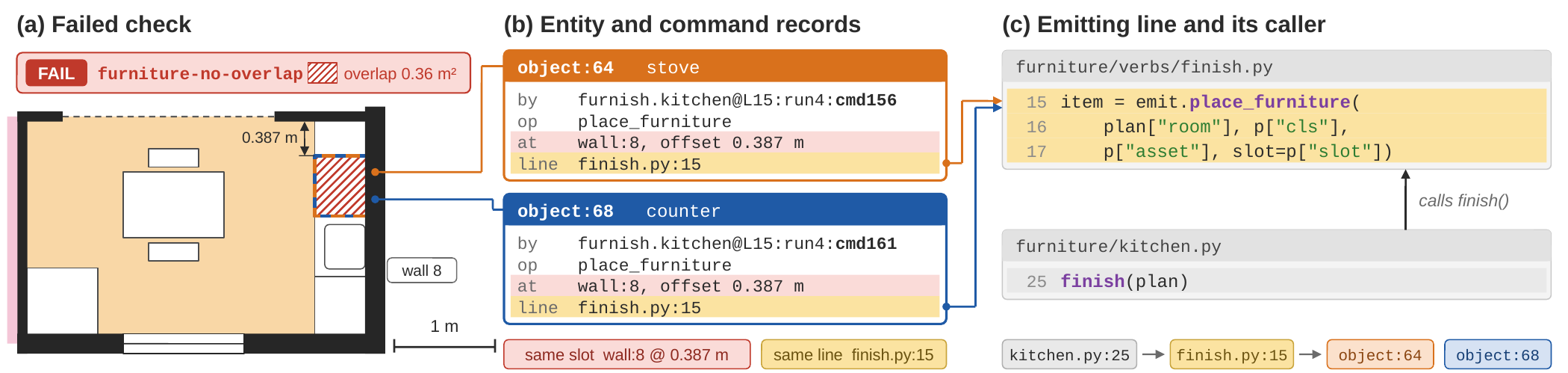}
  \caption{\textbf{From a failed check to the code available for revision.}
    (a) A stove and counter overlap.
    (b) Records for the two elements identify their placement commands.
    (c) The commands identify the source line that issued them and the kitchen function that supplied the placement arguments.
    Colors follow the two objects across panels.}
  \label{fig:method_provenance}
\end{figure*}

\paragraph{Procedural modeling.}
Procedural models encode a content class as a program, from L-systems for plants~\cite{prusinkiewicz1990abop} and shape grammars~\cite{stiny1972shape} to cities and buildings~\cite{parish2001procedural,wonka2003instant,mueller2006procedural,kelly2011interactive,merrell2023example}; see~\cite{smelik2014survey} for a survey.
For interiors, rule-based methods subdivide a footprint into rooms~\cite{marson2010floorplans,lopes2010constrained} or arrange furniture according to interior design guidelines~\cite{merrell2011interactive,yu2011makeithome}, while example-based models and stochastic grammars capture how objects co-occur~\cite{fisher2012example,qi2018human}.
At scale, ProcTHOR~\cite{deitke2022procthor} samples interactive houses for embodied-agent training, and Infinigen~\cite{raistrick2023infinigen} and Infinigen Indoors~\cite{raistrick2024infinigen} combine procedural assets with a constraint language and solver for arranging them.
Writing such rules and choosing their distributions requires considerable expertise, so prior work controls or inverts given procedural models~\cite{talton2011metropolis,ritchie2015controlling,stava2010inverse,vanegas2012inverse,martinovic2013bayesian,wu2014inverse,nishida2016interactive}.
Closest in spirit, Meta-Sim~\cite{kar2019metasim} learns to adjust the attributes of scenes sampled from a probabilistic scene grammar, and Meta-Sim2~\cite{devaranjan2020metasim2} additionally learns the scene structure, \ie, which rule expansions to sample, with reinforcement learning, such that generated scenes match the feature distribution of real images.
In all these works, the grammar itself is written by human experts; we instead acquire and revise the generator program from reference data.

\paragraph{Program synthesis.}
Visual program inference recovers a program that explains a given target, \eg, constructive solid geometry~\cite{sharma2018csgnet}, drawings~\cite{ellis2018learning} or 3D shapes~\cite{tian2019learning,jones2020shapeassembly}, and library learning extracts reusable abstractions~\cite{ellis2021dreamcoder,jones2021shapemod}.
Language models trained on code~\cite{chen2021codex,li2022alphacode} write indoor scene programs~\cite{aguinakang2024openuniverse,gumin2025procedural}, Blender code~\cite{hu2024scenecraft}, parameters of procedural generators~\cite{sun20253dgpt} or rules for procedural buildings~\cite{huang2025buildingblock}, often with feedback from a vision-language model~\cite{huang2024blenderalchemy,yin2026viga}; beyond self-critique~\cite{madaan2023selfrefine,shinn2023reflexion}, an automated evaluator enables evolutionary search over programs~\cite{lehman2022elm,romeraparedes2024funsearch,novikov2025alphaevolve} and reward functions~\cite{ma2024eureka}.
We instead seek a probabilistic program whose output is a dataset of 3D scenes, for which no single known evaluator exists. Related to methods that describe differences between text~\cite{zhong2022describing} or image sets~\cite{dunlap2024describing}, our Discriminator Agent therefore proposes tests of distributional discrepancies together with constraint checks for individual scenes.
Adversarial training dates back to artificial curiosity~\cite{schmidhuber1990making,schmidhuber1991possibility,schmidhuber2020gans} and predictability minimization~\cite{schmidhuber1992predictability} and was popularized by GANs~\cite{goodfellow2014gan}; unlike a GAN discriminator, ours provides inspectable criticism instead of gradients.

\paragraph{Architectural floorplan generation.}
Early approaches cast layout design as optimization, using learned architectural programs with stochastic search~\cite{merrell2010computer}, constrained exploration of layout spaces~\cite{bao2013generating}, deformable templates~\cite{peng2014computing}, or mixed-integer programming~\cite{wu2018miqp}.
Datasets of vectorized residential plans and building complexes~\cite{liu2017raster,kalervo2019cubicasa5k,wu2019data,standfest2022swissdwellings,vanengelenburg2024msd} shifted the field toward learned generators conditioned on boundaries and room graphs~\cite{wu2019data,hu2020graph2plan,nauata2020house,nauata2021house,para2021generative,sun2022wallplan,he2022iplan}, diffusion models~\cite{shabani2023housediffusion,wang2025diffplanner,lian2026cgfloor}, and language~\cite{leng2023tell2design,qin2026tokenization}; Building-GAN~\cite{chang2021buildinggan} extends graph conditioning to multi-story volumetric design.
Learned room-arrangement methods~\cite{wang2018deep,wang2019planit,paschalidou2021atiss,tang2024diffuscene}, trained on synthetic furnished rooms~\cite{fu20213dfront}, address the interior contents.

\paragraph{Agentic scene synthesis.}
Language models can produce layouts directly as structured text~\cite{feng2023layoutgpt} or plan rooms and object placements that are then resolved under spatial constraints~\cite{yang2024holodeck,fu2024anyhome,littlefair2025flairgpt,sun2025layoutvlm}, optionally guided by image generation~\cite{wang2024architect}.
Following multi-agent frameworks that split complex tasks across specialized roles~\cite{hong2024metagpt,qian2024chatdev,wu2024autogen} or let agents critique each other~\cite{du2024debate}, conversing agents have been used to furnish rooms~\cite{celen2024idesign}, build 3D scenes~\cite{yang2025sceneweaver} and design residential floorplans~\cite{zeng2025card}.
Closest to our setting, SceneSmith~\cite{pfaff2026scenesmith} constructs simulation-ready indoor scenes through architectural layout, furniture placement, and small-object population, with designer, critic, and orchestrator agents refining each scene.
The main difference from our work is the unit being revised. We create a reusable stochastic generator that produces a dataset of 3D scenes with high diversity, while SceneSmith generates individual 3D scenes.
Consequently, our language-model cost is paid while learning the program, not for every generated 3D scene.
\section{Method}
\label{sec:method}

\newcounter{apmalgorithm}
\newenvironment{apmalgorithmbox}[2]{%
  \begin{figure}[!ht]
  \begin{minipage}{\linewidth}
  \refstepcounter{apmalgorithm}\label{#1}%
  \small\hrule\smallskip
  \textbf{Algorithm~\theapmalgorithm: #2}\par\smallskip
}{%
  \smallskip\hrule
  \end{minipage}
  \end{figure}
}
\newenvironment{apmalgorithmsteps}{%
  \begin{list}{\arabic{enumi}.}{%
    \usecounter{enumi}%
    \setlength{\leftmargin}{1.5em}%
    \setlength{\labelwidth}{1em}%
    \setlength{\labelsep}{0.5em}%
    \setlength{\itemsep}{1pt}%
    \setlength{\parsep}{0pt}%
    \setlength{\topsep}{2pt}%
  }
}{\end{list}}


Our aim is to develop an agentic system (\cref{fig:teaser}),  whose output is a reusable probabilistic program, a \emph{G-program}, that can be used to generate many 3D scenes depicting architectural layouts including furnishing and decoration.
Examples range from individual kitchens and bathrooms to complete apartments and multi-story buildings. What sets our method apart from earlier adversarial approaches~\cite{schmidhuber1990making,schmidhuber1991possibility,schmidhuber1992predictability,schmidhuber2020gans,goodfellow2014gan} is that coding agents revise procedural generator and discriminator programs, with scene-to-code traces guiding revisions to the generator. 

We start with four inputs: (1) statistics describing how properties of the intended buildings vary, obtained, for example, from one or more reference datasets; (2) a set of \emph{constraints} that every generated scene must satisfy, such as requirements from architectural guidelines or municipal codes; (3) additional guidance specifying the class of scenes to be modeled; and (4) the selected 3D asset catalog, which provides the furniture and decorative objects. Together, these inputs form the user's \emph{task specification}. The user may also provide annotated scenes as examples of desired or unwanted configurations. The goal is for each generated scene to satisfy the constraints and for a scene collection (a \emph{batch}) to match the specified statistics. 

The output of our system is the probabilistic program itself. Each time this program is run, it generates a random 3D scene intended to match the task specification. 
We represent each generated scene as a \emph{structured scene description}.
It records the attributes and geometry of rooms, architectural elements, furniture, and decorative objects.
Crucially, it also links related elements to each other and each element to the program instructions that produced it (its \emph{trace}). Together, these links help the Generator Agent trace problems involving multiple elements back to the relevant code (\cref{fig:method_provenance}).


In the rest of this section, we describe the workflow (\cref{sec:workflow}), assessment (\cref{sec:checks}), and tracing results to source code (\cref{sec:provenance}). We then describe orchestration and the construction of a quality score (\cref{sec:orchestrator}). 


\subsection{Workflow}
\label{sec:workflow}
Our framework follows the generator--discriminator structure of a generative adversarial network (GAN)~\cite{goodfellow2014gan}, with executable programs in place of neural networks.
The programs' source code plays the role of model parameters, while the two LLM-based agents act as optimizers by editing that code. 

Following the GAN analogy, we need a way to evaluate the G-program; in our case we don't particularly care about the nature of its code itself, but rather we focus on examining the quality of its outputs. At development version $t$, we run the G-program $G^{(t)}$ with different random seeds to produce a batch $B^{(t)}=\{s_1,\ldots,s_n\}$. The D-program $D^{(t)}$ then checks the batch using statistical comparisons to the reference data and checks against constraints such as those from the task specification. The results are saved to a database, and then the Generator Agent uses queries against that database as feedback to revise the G-program. The D-program, of course, may also need to be updated. Its checks may miss a problem or incorrectly flag a plausible scene. When this happens, the discriminator agent revises the D-program. Orchestration determines whether this has occurred (\cref{sec:orchestrator}). Each development version records the current pair of programs, including either program when it is unchanged. 

Once developed, the G-program can generate additional scenes without further LLM calls. The cost of agentic development is therefore amortized across the scenes it produces, with no per-scene LLM API cost. Because generation runs the saved program, it also does not depend on later updates to the LLMs used to develop it.


\providecommand{\na}{\textcolor{gray}{n/a}}
\providecommand{\up}{$\uparrow$}
\providecommand{\dn}{$\downarrow$}
\begin{table*}[t]
  \centering
    \caption{\textbf{Complete-scene generation.} (a)~Scenes released by their authors; (b)~all methods on the same 1{,}000 briefs. Metrics as in \cref{sec:evaluation}; means over homes, per column \rankkey{rankfirst}{\textbf{best}}, \rankkey{ranksecond}{second} and \rankkey{rankthird}{third best}. $^\ast$Reported by the authors. $^\dagger$Requires local GPUs or paid API calls, not included.}
  \label{tab:scenes}
  \small
  \setlength{\tabcolsep}{4pt}
  \resizebox{\textwidth}{!}{%
  \begin{tabular}{lr|cc|c|ccc|cccc|r|cc}
    \toprule
    & & \multicolumn{2}{c|}{Floor plan} & Constraints & \multicolumn{3}{c|}{Adherence} & \multicolumn{4}{c|}{Plausibility} & & \multicolumn{2}{c}{Cost / home} \\
    Method & $N$ & Valid\,\%\up & Reach\,\%\up & Hard constr.\,\%\up & Types\,\%\up & Graph\,\%\up & Area\dn & ACC\,\%\up & NAV\,\%\up & COL\,\%\dn & OOB\,\%\dn & \#Obj & \$\dn & min\dn \\
    \midrule
    \multicolumn{15}{l}{\emph{(a) Released scenes}} \\
    SceneSmith & 31 & \secondbest{90.3} & \best{100.0} & \secondbest{77.5} & -- & -- & -- & 94.9 & 97.3 & 8.12 & 0.20 & 214.1 & 13.98$^\ast$ & 206$^\ast$ \\ 
    MolmoSpaces & 636 & \thirdbest{84.4} & \best{100.0} & \thirdbest{67.2} & -- & -- & -- & 96.0 & \thirdbest{99.6} & 12.57 & 0.23 & 159.5 & \na & \na \\
    \midrule
    \multicolumn{15}{l}{\emph{(b) Same briefs (1{,}000 single-story homes)}} \\
    ProcTHOR & 1000 & 33.4 & \thirdbest{96.1} & 65.9 & \best{100.0} & \thirdbest{1.7} & \secondbest{2.1\%} & \secondbest{98.7} & \best{99.8} & \thirdbest{3.69} & \secondbest{0.02} & 73.4 & 0\rlap{$^\dagger$} & \best{0.33} \\
    Infinigen Indoors & 1000 & 73.9 & \secondbest{98.8} & 60.8 & \best{100.0} & \secondbest{97.0} & \thirdbest{2.6\%} & 90.8 & 96.3 & 6.39 & 1.51 & 143.9 & 0 & 178.4 \\
    Holodeck (Qwen3.8-27B) & 826 & 76.5 & 89.7 & 64.3 & \thirdbest{97.8} & \na & 57.0\% & 97.7 & 99.2 & \secondbest{2.97} & \thirdbest{0.10} & 87.3 & 0\rlap{$^\dagger$} & \thirdbest{17.2} \\
    Holodeck (Gemma-4-31B) & 778 & 82.0 & 93.1 & 57.8 & \secondbest{99.7} & \na & 29.7\% & \thirdbest{97.8} & 99.2 & 5.25 & 0.14 & 93.2 & 0\rlap{$^\dagger$} & 22.4 \\
    \apm (ours) & 1000 & \best{99.1} & \best{100.0} & \best{99.5} & \best{100.0} & \best{99.7} & \best{0.0\%} & \best{99.6} & \secondbest{99.7} & \best{0.36} & \best{0.00} & 88.8 & 0 & \secondbest{0.39} \\
    \bottomrule
  \end{tabular}}
\end{table*}


\subsection{Assessing Generated Scenes} 
\label{sec:checks}
To assess the generated scenes, $D^{(t)}$ applies a set of \emph{checks} and a set of \emph{measurements} to each scene in $B^{(t)}$. We denote these sets by \[ \mathcal{K}^{(t)}=\{c_i^{(t)}\}_{i=1}^{K}, \qquad \mathcal{M}^{(t)}=\{m_j^{(t)}\}_{j=1}^{L}. \]
A \emph{check} tests a condition, such as whether furniture overlaps a doorway; a \emph{measurement} computes a property, such as room area. A check $c_i^{(t)}(s)$ returns \emph{pass} if its condition is satisfied, \emph{fail} if it is violated, or \emph{unknown} if the necessary information is missing.
Checks determine whether individual scenes satisfy the constraints, while results across a batch can be compared with statistics from reference data. 

\paragraph{Constraints.} 
\emph{Constraints} are a subset of the checks that every generated scene is required to satisfy, such as room access and the presence of essential fixtures.
Constraints may also be drawn from reference data.
Other than constraints, checks do not need to pass for every scene. Instead, we compare how often their conditions hold across the batch with reference data. For example, a check for a bathtub can be used to compare the fraction of generated bathrooms with bathtubs to the corresponding fraction in the reference data.

\paragraph{Statistical comparisons.} 
Passing constraints in individual scenes does not establish whether a G-program produces the variation found in real homes.
The D-program therefore compares distributions of properties such as room areas, room counts, and object presence across generated scenes and reference homes. For a measurement $m_j^{(t)}$, the values $(m_j^{(t)}(s_1),\ldots,m_j^{(t)}(s_n))$ describe the measured property across the generated batch.
We compare summaries or distributions of these values with reference data, using relative differences for rates or means, Kolmogorov--Smirnov distance for numerical distributions, and total variation distance for categorical distributions. To make these different comparisons usable together, each produces a discrepancy $d_c\in[0,1]$, calibrated so that zero indicates agreement and larger values indicate greater differences. These comparisons can expose a generator that repeatedly produces a narrow range of otherwise acceptable scenes, analogous to mode collapse in GANs.

Statistical comparisons require reference data for the properties being assessed. However, no single dataset may provide all the properties relevant to a task, so the task specification can draw on multiple sources. For example, it may use layout statistics from Swiss Dwellings~\cite{standfest2022swissdwellings}, ZInD~\cite{Cruz_2021_CVPR}, or CubiCasa5K~\cite{kalervo2019cubicasa5k} together with furnishing and decoration statistics from 3D-FRONT~\cite{fu20213dfront} and HSSD~\cite{khanna2024habitat}.


\begin{figure*}[t]
  \centering
  \includegraphics[width=\textwidth]{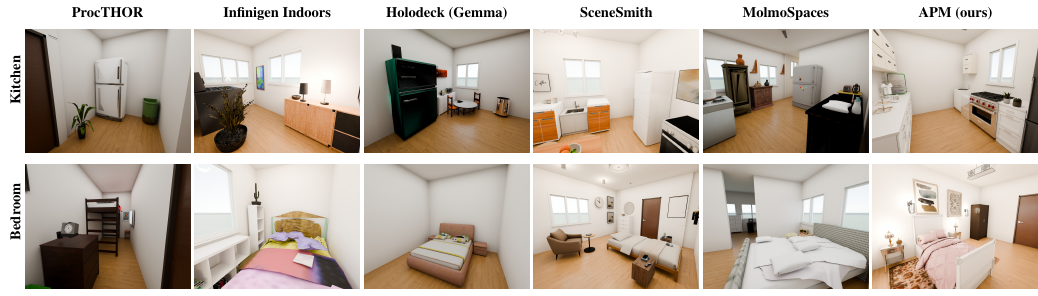}
  \caption{\textbf{Kitchens and bedrooms of every method}, rendered with the same lighting and camera rules. ProcTHOR, Infinigen Indoors, Holodeck and \apm share the brief within a row; SceneSmith and MolmoSpaces show scenes released by their authors.}
  \label{fig:rooms}
\end{figure*}

\subsection{Tracing Results to Source Code}
\label{sec:provenance}
Once an assessment reveals a problem, the Generator Agent must determine which part of the G-program to revise. To do so, the scene’s provenance links are followed to trace the problem back to the lines of code that produced the relevant elements. The process is indicated by the green arrows in~\cref{fig:teaser}. \Cref{fig:method_provenance} shows a more detailed example of a kitchen in which a failed overlap check identifies the stove and counter in (a).
The database records for these elements lead to their placement commands in (b), then to the source line that issued each command in (c).
The records also identify the calling function that supplied the placement arguments.

This information is used to identify the lines of code in the G-program that are most strongly associated with constraint failures and statistical discrepancies. Formally, let $\mathcal{L}_c(s)$ be the source locations reached by tracing check $c$'s failures in scene $s$. For each location $\ell$, we count how many scenes have a failure associated with it: \[ N_c(\ell)=\sum_{s\in B^{(t)}} \mathbf{1}\!\left[\ell\in\mathcal{L}_c(s)\right]. \] The Generator Agent uses these counts to identify code associated with repeated failures. For example, if doorway-blocking failures point to furniture placement, the agent can revise that code to preserve more clearance around doors. It then runs the revised G-program on a new batch to check whether the change reduced the failures.
A caveat is that the trace records code, not data flow (such as how arguments and parameters are computed). To check data flow, the Generator Agent is instructed to rerun selected failed examples with their recorded seeds. Provenance thus plays a credit-assignment role analogous to a gradient, linking feedback to instructions worth changing without computing derivatives. 


\subsection{Orchestrating Program Development}
\label{sec:orchestrator} 

In a GAN, real/fake labels provide feedback for improving the discriminator.
Here, we introduce an Orchestrator Agent (\cref{fig:teaser}) to determine when the D-program needs revision. To do so, it considers the task specification, recorded results, selected visual reviews, and user feedback. When the assessment identifies a problem with the generated scenes, it directs the Generator Agent to revise the G-program; when the assessment itself is flawed, it directs the Discriminator Agent to revise the D-program. The agents can work asynchronously. The user decides when to stop.


\subsubsection{Revising the D-program}
The D-program applies executable checks and measurements across whole batches without per-scene LLM calls.
The orchestrator uses the stored results and inspection by vision-capable agents of selected renderings (examples are shown in~\cref{fig:teaser}) to assess whether those checks capture the relevant problems.
For example, a scene may pass the existing checks while containing a problem those checks overlook, such as `too many wardrobes in one bedroom'. If the orchestrator finds issues, it instructs the Discriminator-Agent to revise the D-program. The same scenes that exposed the problem are used to test the new checks. Examples of issues identified this way during system evaluation are documented in~\cref{tab:objective_trace_apartments,tab:objective_trace_us} of the supplementary material.
Together, revisions to the G- and D-programs aim to produce a higher-quality generator. To measure that improvement, we need criteria for judging its output. But those criteria are not all known at the outset: some emerge as the agents discover shortcomings in generated scenes. We use the D-program’s checks and measurements to compute a quality score.



\subsubsection{Quality score}
\label{sec:quality_score}
\begin{figure*}[t]
  \centering
  \includegraphics{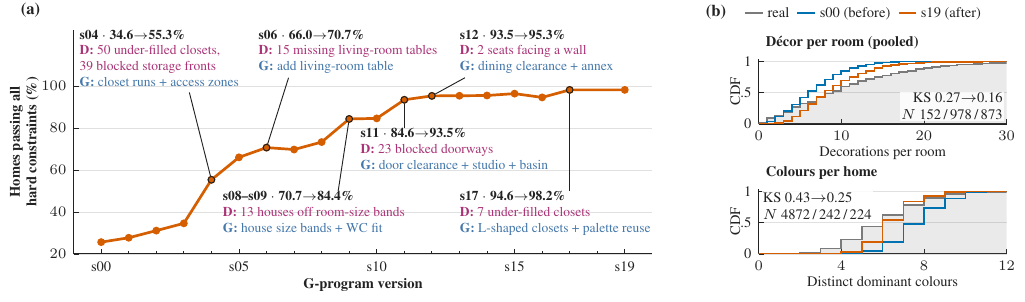}
  \caption{\textbf{Development of the generator.}
    (a)~Generated homes passing all hard constraints across 19 revisions. Each label gives the main problem the D-program found (D) and how the G-program was changed to fix it (G).
    (b)~D\'ecor per room (living rooms, kitchens and bedrooms pooled) and colours per home, s00 (blue) and s19 (orange) against real data (grey), as cumulative distributions (CDF: share of rooms or homes with at most the given count); KS: Kolmogorov--Smirnov distance to the real data, $N$: rooms or homes (real/s00/s19).}
  \label{fig:trajectory}
\end{figure*}

To decide whether a revision improves the G-program, the Orchestrator must evaluate a batch of 3D scenes.
We therefore combine the D-program’s assessment results into a quality score,  $Q(G^{(t)};D^{(t)})$. 
These assessment results summarize the outcomes of checks in $\mathcal{K}^{(t)}$ and measurements in $\mathcal{M}^{(t)}$ across the batch. The score also includes \emph{generation yield} and the \emph{hard-constraint pass rate}, which are the fraction of attempted seeds that produce a completed scene and the fraction of completed scenes that pass every hard-constraint check, respectively. Both have a target of one.
For each of the included comparisons, the discrepancy $d_c$
measures the difference from a reference statistic or design target. We compute the quality score as
\begin{equation}
    Q(G^{(t)};D^{(t)})=100\sum_{c=1}^{C}\alpha_c (1-d_c),
        \label{eq:apm_quality_score}
\end{equation}
where $C$ is the number of comparisons and $\alpha_c\geq 0$ is the relative weight of comparison $c$, chosen to reflect its importance for the task specification, and $\sum_c\alpha_c=1$. 
The Discriminator Agent proposes the numerical weights $\alpha_c$. 
The supplementary material specifies treatment of unavailable measurements (\cref{sec:supp_method_feedback}).
 As checks change, stored scenes from earlier G-programs can be reassessed under the same criteria, allowing their scores to be compared (\cref{fig:trajectory}).


\subsubsection{Selecting generator revisions}
Each proposed G-program revision, $G_{\mathrm{cand}}$, is a \emph{candidate}; the previously accepted version it aims to improve, $G_{\mathrm{parent}}$, is its \emph{parent}.
To distinguish improvement from sampling variation, we evaluate both on the same seeds, under matched execution settings, and with the same D-program configuration.
We measure the score difference as $\Delta Q=Q(G_{\mathrm{cand}};D^{(t)})-Q(G_{\mathrm{parent}};D^{(t)})$. To estimate how much this difference depends on the sampled scenes, we resample the matched seeds 400 times to obtain a 95\% confidence interval for $\Delta Q$ (a paired bootstrap). If this interval lies entirely above zero and the regression checks pass, the Orchestrator reviews selected renderings with a vision-capable agent before deciding whether to accept the candidate.
An accepted candidate becomes the current G-program; a rejected candidate leaves it unchanged. 
A candidate that scores well but produces visibly flawed scenes can also reveal a missing check, prompting the Orchestrator to reassess the D-program.
\Cref{sec:supp_method_selection} of the supplementary material describes the regression tests, including Holm's correction~\cite{holm1979simple}, and other acceptance criteria. When the Orchestrator reports that the user’s targets have been met, the user may opt to provide further guidance, or the user may stop development and accept the G-program.


\section{Experiments}
\label{sec:evaluation}
We compare complete homes generated by \apm with multiple competing methods.
We measure room-layout validity, adherence to the brief, usability of the furnished rooms, physical plausibility of the object placement and the cost of sampling (\cref{tab:scenes}), and visual preference in pairwise comparisons (\cref{tab:userstudy}).
The room-layout stage alone is evaluated in the supplement\ifdefined\includeapmsupplement{} (\cref{sec:layout-benchmark}, \cref{tab:layouts})\fi.

\paragraph{Briefs.}
We generated 1{,}000 single-story homes with \apm. For each of them, \apm first samples a brief, then generates a home from it; the briefs are given to every method in block~(b) of \cref{tab:scenes}.
For one home, the brief lists the rooms with their types, a target net area per room, and the room graph: which rooms are connected by a door or an opening, and which room holds the entrance.
Each brief has one of three layout styles, chosen by the seed: 438 in the style of Swiss apartments, 355 Finnish, and 207 US.
The room programs of each style are drawn from priors fitted to a matching reference dataset: Swiss Dwellings~\cite{standfest2022swissdwellings}, CubiCasa5k~\cite{kalervo2019cubicasa5k} and ZInD~\cite{Cruz_2021_CVPR}.

\paragraph{Baselines.} \emph{ProcTHOR}~\cite{deitke2022procthor} builds a floor plan from our full brief and furnishes it with AI2-THOR assets.
\emph{Infinigen Indoors}~\cite{raistrick2024infinigen} solves floor plan and object placement procedurally under constraints; it receives the same full brief, and its floor plan is restricted to it.
\emph{Holodeck}~\cite{yang2024holodeck} plans rooms and objects with a language model from a text query (room counts and total area, without areas per room or the room graph).
We replace its GPT-4 with newer open-weight models, Qwen3.8-27B and Gemma-4-31B.
Holodeck produced no home for many of the larger briefs (\cref{sec:supp_baselines}).
\emph{SceneSmith}~\cite{pfaff2026scenesmith} is an agentic system; we score the 31 house-level scenes its authors released.
\emph{MolmoSpaces} \cite{kim2026molmospaces} released homes generated with an extended Holodeck; we use the 636 homes of its validation split.
Both follow prompts of their own and are not conditioned on our briefs, so block~(a) of \cref{tab:scenes} has no adherence scores.
House-GAN++~\cite{nauata2021house}, HouseDiffusion~\cite{shabani2023housediffusion} and WallPlan~\cite{sun2022wallplan} generate room layouts only, from RPLAN~\cite{wu2019data} inputs, and appear in the room-layout benchmark of the supplement.
Configurations of all baselines are given in the supplement\ifdefined\includeapmsupplement{} (\cref{sec:supp_baselines})\fi.

\paragraph{Hard constraints.}
The hard constraints given to the method, input~(2) of \cref{sec:method}, are 24 checks covering access, required fixtures and furniture, object placement and storage, and room-program plausibility.
Each check returns pass, fail, or unknown for every element it judges; the list is given in the supplement\ifdefined\includeapmsupplement{} (\cref{sec:supp_constraints})\fi.
\emph{Hard constr.} in \cref{tab:scenes} is the \% of checks a home passes, averaged over homes.

\paragraph{Metrics.}
All metrics are computed on each method's own room layout and its objects' oriented boxes.
\emph{Floor plan.} \emph{Valid}: the layout has no overlapping rooms, no holes, no disconnected rooms, and no room below the minimum area or width or above the maximum aspect ratio of its type; the thresholds are loose fit thresholds, the same for every method. \emph{Reach}: every room can be reached from the entrance through doors or openings.
\emph{Adherence} to the brief. \emph{Types}: the room types and counts equal the brief; \emph{Graph}: the door graph equals the specified room graph (not applicable to Holodeck, whose input contains no graph); \emph{Area}: relative error of the net area.
\emph{Plausibility}, following SceneEval~\cite{tam2026sceneeval} and computed on oriented object boxes for every method. \emph{ACC}: floor objects an agent of 0.2\,m radius can reach; \emph{NAV}: free floor connected for that agent, both per room; \emph{COL}: objects whose box intersects another object's box (objects resting on or in another, stacks of one kind, chairs tucked under tables and rugs are not counted); \emph{OOB}: objects outside the rooms.
\emph{\#Obj}: objects per home. \emph{Cost}: paid API cost per home and wall-clock minutes per delivered home of one process, including failed attempts.
Thresholds, exact definitions and the hardware of each method are given in the supplement\ifdefined\includeapmsupplement{} (\cref{sec:supp_scene_metrics})\fi.

\providecommand{\tbd}{\textcolor{gray}{--}}
\begin{table}[t]
  \centering
  \caption{\textbf{Visual preference.} Share of pairwise judgements preferring \apm; 95\% confidence intervals. User study: 35 participants.}
  \label{tab:userstudy}
  \small
  \setlength{\tabcolsep}{4pt}
  \resizebox{\columnwidth}{!}{%
  \begin{tabular}{l|cc@{\hspace{6pt}\vrule width 1.2pt\hspace{6pt}}c}
    \toprule
    & Sonnet 5.5 & GPT-6.1 Sol & User study \\
    \apm vs. & Realism\,\% & Realism\,\% & Realism\,\% \\
    \midrule
    ProcTHOR               & 100.0\,{\scriptsize[96.3, 100.0]} & 97.0\,{\scriptsize[91.5, 99.0]} & 93.3\,{\scriptsize[89.2, 95.9]} \\
    Infinigen Indoors      & 83.0\,{\scriptsize[74.5, 89.1]} & 86.0\,{\scriptsize[77.9, 91.5]} & 70.4\,{\scriptsize[64.0, 76.1]} \\
    Holodeck (Qwen)        & 75.0\,{\scriptsize[65.7, 82.5]} & 76.0\,{\scriptsize[66.8, 83.3]} & 84.4\,{\scriptsize[79.1, 88.6]} \\
    Holodeck (Gemma)       & 73.0\,{\scriptsize[63.6, 80.7]} & 67.0\,{\scriptsize[57.3, 75.4]} & 70.5\,{\scriptsize[64.2, 76.0]} \\
    \bottomrule
  \end{tabular}}
\end{table}
\subsection{Results: complete scenes}
\label{sec:exp_results}
\Cref{tab:scenes}\,(b) compares the methods on the same briefs.
\apm has the most valid floor plans: 99.1\% against 33.4--82.0\% for the baselines.
Its door graphs match the specified room graph in 99.7\% of homes, and its net areas deviate by 0.0\% on average; Infinigen Indoors matches the graph in 97.0\% of homes and ProcTHOR in 1.7\%.
Types is close to 100\% for every method that receives the full brief, since the room list is part of its input; Holodeck, which receives only a text query, reaches 97.8\% and 99.7\%, but its net areas deviate by 57.0\% and 29.7\%.
Every room is reachable in all homes of \apm; the baselines reach 89.7--98.8\%.

The furnished rooms of \apm have the fewest colliding objects (0.36\% against 2.97--6.39\%) and no objects outside the rooms.
Its accessibility is the highest (99.6\%); navigability is similar for all methods except Infinigen Indoors (96.3\%).
\apm places 88.8 objects per home, between ProcTHOR (73.4) and Infinigen Indoors (143.9); a low collision rate is therefore not the result of sparse rooms.
Sampling a home takes 0.39\,min on one CPU core without API cost; Infinigen Indoors takes 178.4\,min per home and Holodeck 17.2--22.4\,min on local GPUs.
The released scenes of block~(a) are not comparable on adherence or cost: they follow other prompts, and SceneSmith's cost and time (\$13.98, 206\,min) are the authors' figures per room-level scene, the only ones they report.
\Cref{fig:rooms} shows a kitchen and a bedroom for every method.
ProcTHOR, Infinigen Indoors, Holodeck and \apm received the same brief in each row, taken from the homes of the perceptual study (\cref{sec:exp_perceptual}); SceneSmith and MolmoSpaces show released scenes.
Complete homes with top views are given in the supplement\ifdefined\includeapmsupplement{} (\cref{fig:qualitative})\fi.

\subsection{Perceptual study}
\label{sec:exp_perceptual}

We compare \apm with each baseline in blinded pairs on 100 shared briefs, each home shown as one top-down view from the same rendering pipeline, and ask \emph{``Which home looks more like a real, lived-in home?''} (A, B or equal).
Two vision-language models, Claude Sonnet~5.5 and GPT-6.1 Sol, and human participants judge the same pairs (\cref{tab:userstudy}; protocol in the supplement\ifdefined\includeapmsupplement{}, \cref{sec:supp_perceptual}\fi).
Both judges prefer \apm against every baseline, in 67--100\% of the pairs; so do the participants (70--93\%, 35 participants).

\subsection{Development trajectory}
\label{sec:exp_trajectory}
\Cref{fig:trajectory}a follows the G-program over 20 versions, s00 to s19, each tested with 300 generation attempts; shares are over the completed homes in the supplement\ifdefined\includeapmsupplement{} (\cref{sec:supp_dev_yield})\fi.
At each version, we report the most frequent failure found by the D-program and the Generator Agent's response;
for example, 23 homes with doorways blocked by furniture at s10 led to larger door clearances in s11, and the share of homes satisfying all hard constraints rose from 84.6\% to 93.5\%.
Over the 20 versions, this share rose from 25.6\% to 98.2\%.
\Cref{fig:trajectory}b shows two statistical comparisons for s00 and s19: decorations per room and distinct colors per home moved toward the reference data (Kolmogorov--Smirnov distance 0.27 to 0.16 and 0.43 to 0.25).

\subsection{Ablations}
\label{sec:exp_ablations}
\providecommand{\tbd}{\textcolor{gray}{--}}
\begin{table}[t]
  \centering
  \small
  \caption{\textbf{Ablations.} Generated homes passing all hard constraints (\%) after 10 revisions; every variant starts from s00 (25.6\%).}
  \label{tab:ablation}
  \begin{tabular}{lcc}
    \toprule
    & 10 revisions & $\Delta$ vs.\ s00 \\
    \midrule
    \apm & \textbf{84.6} & \textbf{+59.0} \\
    w/o traces & 50.0 & +24.4 \\
    w/o D-program feedback & 32.7 & +7.1 \\
    \bottomrule
  \end{tabular}
\end{table}

Both ablations start from the same G-program s00 and run 10 revisions; each was run once.
\emph{Without traces}, the method runs as in our run, with the Discriminator Agent revising the D-program, but the Generator Agent sees only which checks fail and in which homes, not the commands and source lines that produced the failing elements (\cref{sec:provenance}); the share of homes passing all hard constraints reaches 50\% after 10 revisions, against 84.6\% with traces (\cref{tab:ablation}).
\emph{Without D-program feedback}, the Generator Agent receives the full list of requirements from the start (in our run they were added during development), but no check results, statistical comparisons, or traces; it can only inspect generated homes, and reaches 32.7\%.
It fixed the requirements named in the task specification, such as kitchen sets and the colour palette, but not the geometric and functional defects that no requirement names, such as room proportions, storage access, and toilet clearances; these account for most of the gain in our run (s04, s08--s09 and s11 in \cref{fig:trajectory}).
\section{Conclusion}
\label{sec:conclusion}
We propose a new paradigm called \emph{adversarial procedural modeling}. 
The target is a probabilistic program whose repeated execution produces batches of 3D scenes. The program is derived by the interplay of a Generator Agent that revises the probabilistic program and a  Discriminator Agent that writes code to test generated batches of 3D scenes.

\section*{Acknowledgments}
The work is supported by funding from King Abdullah University of Science and Technology (KAUST)—Center of Excellence for Generative AI, under award number 5940, by a gift from Google, and by the Austrian Research Promotion Agency (FFG) under the project ``AUTARK'' (No.~999922723), within the call ``Schlüsseltechnologien im produktionsnahen Umfeld''.
\FloatBarrier
{
    \small
    \bibliographystyle{ieeenat_fullname}
    \bibliography{main}
}

\ifdefined\includeapmsupplement
  \onecolumn
  \renewcommand{\topfraction}{0.95}\renewcommand{\bottomfraction}{0.9}\renewcommand{\textfraction}{0.05}
  \renewcommand{\floatpagefraction}{0.85}\setcounter{topnumber}{4}\setcounter{bottomnumber}{4}\setcounter{totalnumber}{8}
  \appendix
  \begin{center}{\Large\textbf{Supplementary Material}}\end{center}
  \vspace{1em}
\setcounter{figure}{0}
\renewcommand{\thefigure}{S\arabic{figure}}

\setcounter{table}{2}
\renewcommand{\thetable}{S\arabic{table}}
\newcounter{apmsavedtable}

\def\RBversion{next-unified-gl89-51a916e8e6fa}
\def\RBN{987}
\def\RBNRef{18{,}753}
\def\RBNRealAll{46{,}830}
\def\RBrealValid{75.0}
\def\RBoursValid{\textbf{94.4}}
\def\RBprocthorValid{32.9}
\def\RBinfinigenValid{72.1}
\def\RBholoqwenValid{78.0}
\def\RBhologemmaValid{83.8}
\def\RBoursReach{99.0}
\def\RBprocthorReach{95.2}
\def\RBinfinigenReach{\textbf{99.5}}
\def\RBholoqwenReach{86.6}
\def\RBhologemmaReach{91.7}
\def\RBoursTypes{\textbf{100.0}}
\def\RBprocthorTypes{\textbf{100.0}}
\def\RBinfinigenTypes{\textbf{100.0}}
\def\RBholoqwenTypes{98.6}
\def\RBhologemmaTypes{\textbf{100.0}}
\def\RBoursGraph{\textbf{96.1}}
\def\RBprocthorGraph{1.1}
\def\RBinfinigenGraph{60.7}
\def\RBholoqwenGraph{\na}
\def\RBhologemmaGraph{\na}
\def\RBoursArea{\textbf{0.0\%}}
\def\RBprocthorArea{3.1\%}
\def\RBinfinigenArea{7.1\%}
\def\RBholoqwenArea{52.2\%}
\def\RBhologemmaArea{23.3\%}
\def\RBoursDFID{\textbf{45.8}}
\def\RBprocthorDFID{81.4}
\def\RBinfinigenDFID{70.1}
\def\RBholoqwenDFID{92.3}
\def\RBhologemmaDFID{116.7}
\def\RBoursDKID{\textbf{41.5}}
\def\RBprocthorDKID{83.6}
\def\RBinfinigenDKID{78.3}
\def\RBholoqwenDKID{102.1}
\def\RBhologemmaDKID{132.5}
\def\RBoursN{987}
\def\RBprocthorN{987}
\def\RBinfinigenN{987}
\def\RBholoqwenN{854}
\def\RBhologemmaN{745}
\def\RCversion{next-multistory-nav-4648dcf51a12}
\def\RCN{1{,}004}
\def\RCNRef{38{,}152}
\def\RCNRealAll{6{,}103}
\def\RCrealValid{54.9}
\def\RCoursValid{91.2}
\def\RCprocthorValid{67.6}
\def\RCinfinigenValid{78.6}
\def\RCholoqwenValid{\textbf{93.4}}
\def\RChologemmaValid{88.5}
\def\RCoursReach{\textbf{100.0}}
\def\RCprocthorReach{94.3}
\def\RCinfinigenReach{99.3}
\def\RCholoqwenReach{91.6}
\def\RChologemmaReach{97.3}
\def\RCoursTypes{\textbf{100.0}}
\def\RCprocthorTypes{\textbf{100.0}}
\def\RCinfinigenTypes{\textbf{100.0}}
\def\RCholoqwenTypes{48.3}
\def\RChologemmaTypes{81.7}
\def\RCoursGraph{\textbf{100.0}}
\def\RCprocthorGraph{0.1}
\def\RCinfinigenGraph{87.6}
\def\RCholoqwenGraph{\na}
\def\RChologemmaGraph{\na}
\def\RCoursArea{\textbf{0.0\%}}
\def\RCprocthorArea{1.2\%}
\def\RCinfinigenArea{21.4\%}
\def\RCholoqwenArea{30.0\%}
\def\RChologemmaArea{11.7\%}
\def\RCoursDFID{78.4}
\def\RCprocthorDFID{90.8}
\def\RCinfinigenDFID{\textbf{73.1}}
\def\RCholoqwenDFID{98.0}
\def\RChologemmaDFID{139.0}
\def\RCoursDKID{\textbf{65.3}}
\def\RCprocthorDKID{90.8}
\def\RCinfinigenDKID{74.0}
\def\RCholoqwenDKID{118.5}
\def\RChologemmaDKID{173.3}
\def\RCoursN{1004}
\def\RCprocthorN{1004}
\def\RCinfinigenN{1004}
\def\RCholoqwenN{640}
\def\RChologemmaN{715}
\def\RSversion{next-superunified-single-s39-0c15b742b912-1000-swiss}
\def\RSN{438}
\def\RSNRef{8{,}322}
\def\RSNRealAll{46{,}830}
\def\RSrealValid{75.0}
\def\RSoursValid{\best{97.9}}
\def\RSprocthorValid{28.5}
\def\RSinfinigenValid{72.4}
\def\RSholoqwenValid{\thirdbest{76.2}}
\def\RShologemmaValid{\secondbest{84.2}}
\def\RSoursReach{\best{100.0}}
\def\RSprocthorReach{\thirdbest{95.4}}
\def\RSinfinigenReach{\secondbest{98.9}}
\def\RSholoqwenReach{88.3}
\def\RShologemmaReach{91.2}
\def\RSoursTypes{\best{100.0}}
\def\RSprocthorTypes{\best{100.0}}
\def\RSinfinigenTypes{\best{100.0}}
\def\RSholoqwenTypes{\secondbest{97.5}}
\def\RShologemmaTypes{\best{100.0}}
\def\RSoursGraph{\best{100.0}}
\def\RSprocthorGraph{\thirdbest{0.0}}
\def\RSinfinigenGraph{\secondbest{96.6}}
\def\RSholoqwenGraph{\na}
\def\RShologemmaGraph{\na}
\def\RSoursArea{\best{0.0\%}}
\def\RSprocthorArea{\thirdbest{3.1\%}}
\def\RSinfinigenArea{\secondbest{2.8\%}}
\def\RSholoqwenArea{53.1\%}
\def\RShologemmaArea{22.2\%}
\def\RSoursDFID{\best{39.8}}
\def\RSprocthorDFID{\thirdbest{76.0}}
\def\RSinfinigenDFID{\secondbest{65.0}}
\def\RSholoqwenDFID{85.1}
\def\RShologemmaDFID{109.9}
\def\RSoursDKID{\best{39.2}}
\def\RSprocthorDKID{\thirdbest{82.4}}
\def\RSinfinigenDKID{\secondbest{78.0}}
\def\RSholoqwenDKID{98.2}
\def\RShologemmaDKID{131.1}
\def\RSoursN{438}
\def\RSprocthorN{438}
\def\RSinfinigenN{438}
\def\RSholoqwenN{366}
\def\RShologemmaN{323}
\def\RFversion{next-superunified-single-s39-0c15b742b912-1000-finnish}
\def\RFN{355}
\def\RFNRef{6{,}745}
\def\RFNRealAll{6{,}103}
\def\RFrealValid{54.9}
\def\RFoursValid{\best{100.0}}
\def\RFprocthorValid{34.6}
\def\RFinfinigenValid{82.8}
\def\RFholoqwenValid{\secondbest{85.2}}
\def\RFhologemmaValid{\thirdbest{84.1}}
\def\RFoursReach{\best{100.0}}
\def\RFprocthorReach{\secondbest{98.9}}
\def\RFinfinigenReach{\thirdbest{98.6}}
\def\RFholoqwenReach{96.0}
\def\RFhologemmaReach{96.1}
\def\RFoursTypes{\best{100.0}}
\def\RFprocthorTypes{\best{100.0}}
\def\RFinfinigenTypes{\best{100.0}}
\def\RFholoqwenTypes{\secondbest{98.5}}
\def\RFhologemmaTypes{\best{100.0}}
\def\RFoursGraph{\best{100.0}}
\def\RFprocthorGraph{\thirdbest{3.9}}
\def\RFinfinigenGraph{\secondbest{97.7}}
\def\RFholoqwenGraph{\na}
\def\RFhologemmaGraph{\na}
\def\RFoursArea{\best{0.0\%}}
\def\RFprocthorArea{\secondbest{1.8\%}}
\def\RFinfinigenArea{\thirdbest{2.6\%}}
\def\RFholoqwenArea{63.0\%}
\def\RFhologemmaArea{34.9\%}
\def\RFoursDFID{\best{53.4}}
\def\RFprocthorDFID{\thirdbest{58.4}}
\def\RFinfinigenDFID{\secondbest{56.5}}
\def\RFholoqwenDFID{76.3}
\def\RFhologemmaDFID{98.3}
\def\RFoursDKID{\best{62.1}}
\def\RFprocthorDKID{\secondbest{65.5}}
\def\RFinfinigenDKID{\thirdbest{70.4}}
\def\RFholoqwenDKID{96.5}
\def\RFhologemmaDKID{121.1}
\def\RFoursN{355}
\def\RFprocthorN{355}
\def\RFinfinigenN{355}
\def\RFholoqwenN{324}
\def\RFhologemmaN{308}
\def\RUversion{next-superunified-single-s39-0c15b742b912-1000-us}
\def\RUN{207}
\def\RUNRef{3{,}888}
\def\RUNRealAll{2{,}453}
\def\RUrealValid{12.7}
\def\RUoursValid{\best{100.0}}
\def\RUprocthorValid{36.7}
\def\RUinfinigenValid{\thirdbest{61.8}}
\def\RUholoqwenValid{55.8}
\def\RUhologemmaValid{\secondbest{72.8}}
\def\RUoursReach{\best{100.0}}
\def\RUprocthorReach{\thirdbest{91.3}}
\def\RUinfinigenReach{\secondbest{99.0}}
\def\RUholoqwenReach{79.0}
\def\RUhologemmaReach{91.2}
\def\RUoursTypes{\best{100.0}}
\def\RUprocthorTypes{\best{100.0}}
\def\RUinfinigenTypes{\best{100.0}}
\def\RUholoqwenTypes{\thirdbest{97.1}}
\def\RUhologemmaTypes{\secondbest{98.6}}
\def\RUoursGraph{\best{98.6}}
\def\RUprocthorGraph{\thirdbest{0.5}}
\def\RUinfinigenGraph{\secondbest{96.6}}
\def\RUholoqwenGraph{\na}
\def\RUhologemmaGraph{\na}
\def\RUoursArea{\best{0.0\%}}
\def\RUprocthorArea{\best{0.0\%}}
\def\RUinfinigenArea{\secondbest{2.0\%}}
\def\RUholoqwenArea{52.9\%}
\def\RUhologemmaArea{\thirdbest{35.0\%}}
\def\RUoursDFID{51.7}
\def\RUprocthorDFID{\secondbest{28.7}}
\def\RUinfinigenDFID{\best{26.2}}
\def\RUholoqwenDFID{\thirdbest{36.1}}
\def\RUhologemmaDFID{42.4}
\def\RUoursDKID{62.0}
\def\RUprocthorDKID{\best{31.2}}
\def\RUinfinigenDKID{\secondbest{31.7}}
\def\RUholoqwenDKID{\thirdbest{40.7}}
\def\RUhologemmaDKID{46.8}
\def\RUoursN{207}
\def\RUprocthorN{207}
\def\RUinfinigenN{207}
\def\RUholoqwenN{138}
\def\RUhologemmaN{147}
\def\RArealN{1000}
\def\RAhgppN{1000}
\def\RAhdN{996}
\def\RAwallplanN{992}
\def\RAprocthorN{1000}
\def\RAinfinigenN{998}
\def\RAholoqwenN{904}
\def\RAhologemmaN{812}
\def\RAapmbubbleN{903}
\def\RAapmareasN{859}
\def\RArealValid{84.1}
\def\RAhgppValid{44.6}
\def\RAhdValid{60.3}
\def\RAwallplanValid{\best{87.7}}
\def\RAprocthorValid{36.8}
\def\RAinfinigenValid{77.0}
\def\RAholoqwenValid{67.9}
\def\RAhologemmaValid{41.5}
\def\RAapmbubbleValid{\secondbest{85.6}}
\def\RAapmareasValid{\thirdbest{85.2}}
\def\RArealReach{92.1}
\def\RAhgppReach{31.2}
\def\RAhdReach{78.9}
\def\RAwallplanReach{55.1}
\def\RAprocthorReach{\thirdbest{94.5}}
\def\RAinfinigenReach{\secondbest{98.5}}
\def\RAholoqwenReach{48.7}
\def\RAhologemmaReach{66.4}
\def\RAapmbubbleReach{\best{100.0}}
\def\RAapmareasReach{\best{100.0}}
\def\RArealTypes{\na}
\def\RAhgppTypes{\thirdbest{48.1}}
\def\RAhdTypes{\best{100.0}}
\def\RAwallplanTypes{\na}
\def\RAprocthorTypes{\best{100.0}}
\def\RAinfinigenTypes{\best{100.0}}
\def\RAholoqwenTypes{\secondbest{99.6}}
\def\RAhologemmaTypes{\best{100.0}}
\def\RAapmbubbleTypes{\best{100.0}}
\def\RAapmareasTypes{\best{100.0}}
\def\RArealGraph{\na}
\def\RAhgppGraph{21.1}
\def\RAhdGraph{\secondbest{70.7}}
\def\RAwallplanGraph{\na}
\def\RAprocthorGraph{3.7}
\def\RAinfinigenGraph{\thirdbest{55.6}}
\def\RAholoqwenGraph{6.2}
\def\RAhologemmaGraph{20.6}
\def\RAapmbubbleGraph{\best{100.0}}
\def\RAapmareasGraph{\best{100.0}}
\def\RArealArea{\na}
\def\RAhgppArea{\na}
\def\RAhdArea{\na}
\def\RAwallplanArea{\na}
\def\RAprocthorArea{\secondbest{3.9\%}}
\def\RAinfinigenArea{\thirdbest{6.3\%}}
\def\RAholoqwenArea{25.6\%}
\def\RAhologemmaArea{9.2\%}
\def\RAapmbubbleArea{\na}
\def\RAapmareasArea{\best{0.0\%}}
\def\RArealDFID{0}
\def\RAhgppDFID{20.4}
\def\RAhdDFID{\secondbest{5.0}}
\def\RAwallplanDFID{\best{1.7}}
\def\RAprocthorDFID{\thirdbest{14.0}}
\def\RAinfinigenDFID{21.1}
\def\RAholoqwenDFID{17.2}
\def\RAhologemmaDFID{25.0}
\def\RAapmbubbleDFID{28.0}
\def\RAapmareasDFID{24.1}
\def\RArealDKID{0}
\def\RAhgppDKID{18.8}
\def\RAhdDKID{\secondbest{3.9}}
\def\RAwallplanDKID{\best{1.4}}
\def\RAprocthorDKID{\thirdbest{16.2}}
\def\RAinfinigenDKID{26.3}
\def\RAholoqwenDKID{19.8}
\def\RAhologemmaDKID{33.6}
\def\RAapmbubbleDKID{32.6}
\def\RAapmareasDKID{26.6}

\noindent\textbf{Overview.}
In the main paper, we showed that coding agents can learn a reusable procedural program that generates complete, furnished homes without calling a language model at sampling time.
The learned program generated homes about 450 times as fast as Infinigen Indoors.
In a blinded user study, participants judged its homes more realistic in 70\% of comparisons with Infinigen Indoors (not counting ties).
Here we take a closer look at the layouts and furnishings behind those judgments and at how the generator improved over successive revisions.

This supplement is organized as follows.
We start with additional qualitative examples in \cref{fig:qualitative}, showing how different methods lay out and furnish homes for the same four briefs.
We then measure how often layouts satisfy the design constraints (\cref{sec:supp_results}) and report the user study comparing the realism of the furnished homes (\cref{sec:supp_perceptual}).
To understand how the generator arrived at these results, we follow its development and show how the scoring objectives evolved (\cref{sec:supp_development}).
We close by describing how we evaluated the homes (\cref{sec:supp_eval}) and how to reproduce the generation process (\cref{sec:supp_implementation}).

\ifdefined\includeapmsupplement\else
\noindent\textbf{Supplement references in the main paper.}
In the main paper, several items are referred to as ``in the supplement'' without a section number; here we connect those references to specific sections.
For the overhead views mentioned just after the discussion of Fig.~4 in Sec.~4.1 of the main paper, see \cref{fig:qualitative}.
The room-layout benchmark introduced in Sec.~4 of the main paper is reported in \cref{tab:layouts}.

Three passages in the main paper refer to the supplement by section number.
In the discussion of the quality score (Sec.~3.4.2), the pointer to supplement Sec.~2 concerns the treatment of unavailable measurements (\cref{sec:supp_method_feedback}).
The next subsection (Sec.~3.4.3) points to supplement Sec.~4 for the regression tests used to accept a generator revision (\cref{sec:supp_method_selection}).
In the baseline comparison (Sec.~4), the pointer to supplement Sec.~6.1 concerns the Holodeck configuration and the briefs for which it produced no home (\cref{sec:supp_baselines}).
\fi

\section{Additional results and comparisons}
\label{sec:supp_results}
\subsection{Qualitative comparison}
\label{sec:supp_qualitative}
 
This section expands on the qualitative evaluation beyond what could be shown in \ifdefined\includeapmsupplement\cref{fig:rooms}\else Fig.~4\fi{} of the main paper. 
Each row of \cref{fig:qualitative} follows one design brief across four methods. 
The first row illustrates the types of issue that can occur.
ProcTHOR's living room is long and narrow, with a sofa facing a blank wall.
Objects near the door also appear to float.
Infinigen Indoors places a sofa away from the wall, constricting the space behind it.
In the Holodeck plan, the L-shaped area at the left appears inaccessible from the rest of the home.
In \apm{}, the living room has space to group seats around a shared coffee table.

The automatically generated checks in our D-program can identify problems with room usability that are easy to miss in a rendering.
In the second row of \cref{fig:qualitative}, the D-program finds no sink in any of the three baseline kitchens, while the \apm{} kitchen includes one.
In the fourth row, Holodeck places a laundry hamper just 13\,cm in front of the toilet, leaving little room to use it.

The \apm{} home in the first row also leaves room for improvement: the light above the sofa is placed asymmetrically, the toilet's position relative to the bathtub could be improved, and the dining room is far from the kitchen.

Such observations can guide further development; however, taken together with the quantitative and perceptual results in the main paper, these comparisons show that \apm{} improves layout quality (while generating homes about 450 times faster than Infinigen Indoors).

\begin{figure*}[!htbp]
  \centering
  \includegraphics[width=\textwidth,height=0.95\textheight,keepaspectratio]{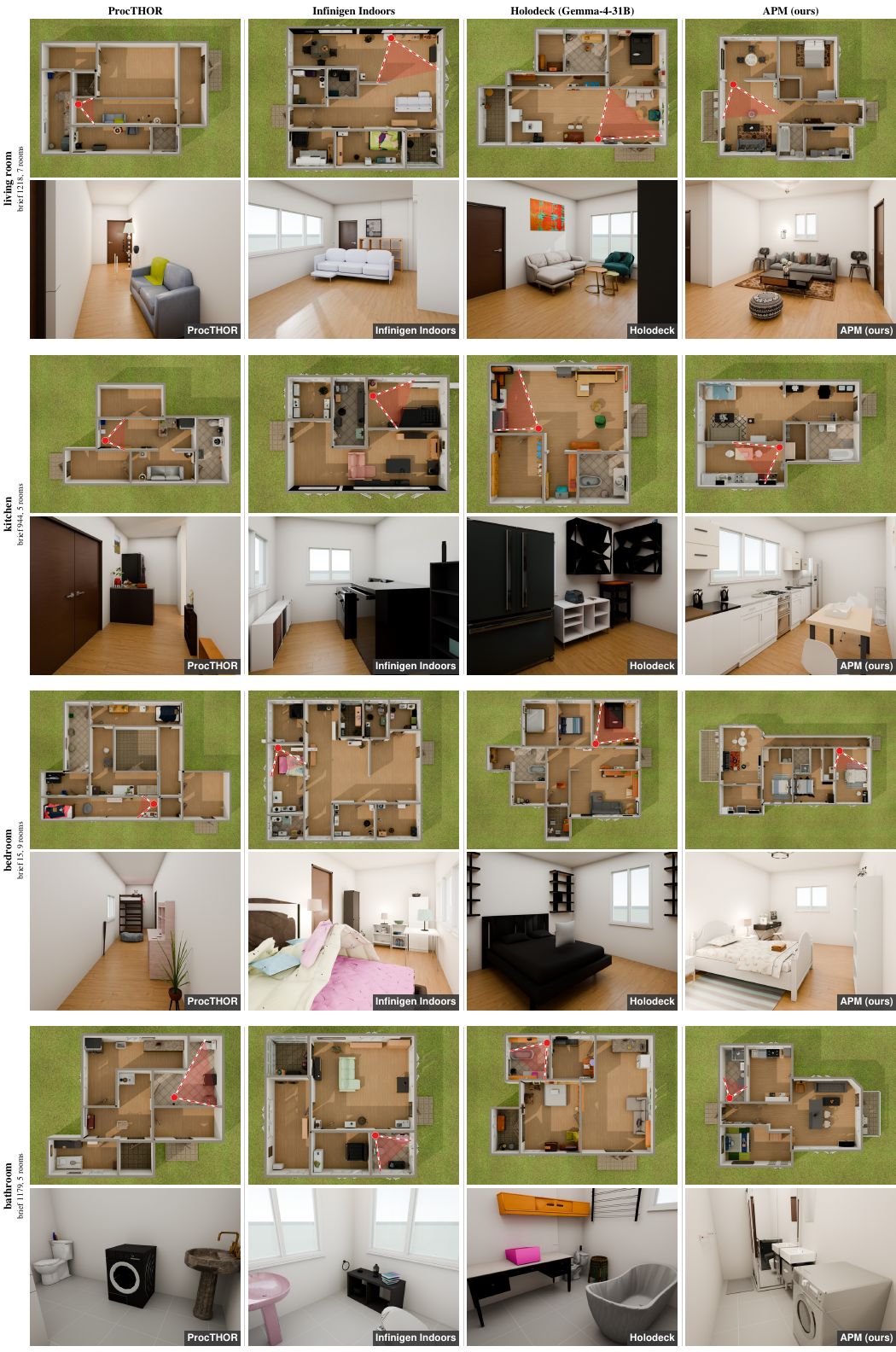}
  \caption{\textbf{The same briefs produce different furniture arrangements.}
In the living-room row, \apm{} groups the seating around a coffee table.
The overhead view locates that group in the room; the red wedge marks the position and FOV of the corresponding interior camera.}
  \label{fig:qualitative}
\end{figure*}

\subsection{Room-layout results}
The G-program we developed to generate complete furnished homes (Sec.~4 of the main paper) can also lay out a prescribed set of rooms (as in HouseDiffusion~\cite{shabani2023housediffusion} and House-GAN++~\cite{nauata2021house}), even though that is not the task our method was designed for.
The G-program normally chooses the rooms and their door connections before laying them out (\cref{sec:supp_method_reproducibility}).
When we instead supply those choices from 1{,}000 apartments from the RPLAN dataset, it produces valid layouts for more apartments than either dedicated floor-plan generator, HouseDiffusion or House-GAN++ (\cref{tab:layouts}, block~(a); examples in \cref{fig:layouts_rplan}).

A second version of the layout problem also gives a target area for each room, the input ProcTHOR, Infinigen Indoors and Holodeck are designed for.
Here \apm{} is the only method that realizes the requested door connections in every layout: each of its 859 layouts has exactly the requested door graph, against 55.6\% of the layouts of Infinigen Indoors, 3.7\% of ProcTHOR's and at most 20.6\% of Holodeck's.
Its rooms also match their target areas better (mean error below 0.1\%, against 3.9--25.6\% for the other methods), the door graph connects every room to the entrance, and 732 of its 859 returned layouts are valid (85.2\%), more than for any of these methods.
The G-program discards the remaining 141 inputs during construction (\cref{sec:supp_dev_yield}); counted over all 1{,}000 apartments, 73.2\% receive a valid layout. This is close to Infinigen Indoors (76.8\%) and twice ProcTHOR (36.8\%).
\Cref{fig:layouts_briefs} shows layouts of all methods for briefs of the 1{,}000 single-story homes (blocks~(b--d)).

\providecommand{\tbd}{\textcolor{gray}{--}}

\def\SAscenesmithObj{214.1}
\def\SAmolmospacesObj{159.5}
\def\SAscenesmithValid{\textbf{90.3}}
\def\SAmolmospacesValid{84.4}
\def\SAscenesmithReach{\textbf{100.0}}
\def\SAmolmospacesReach{\textbf{100.0}}
\def\SAscenesmithACC{94.9}
\def\SAmolmospacesACC{\textbf{96.0}}
\def\SAscenesmithNAV{97.3}
\def\SAmolmospacesNAV{\textbf{99.6}}
\def\SAscenesmithCOL{\textbf{8.12}}
\def\SAmolmospacesCOL{12.57}
\def\SAscenesmithOOB{\textbf{0.20}}
\def\SAmolmospacesOOB{0.23}
\def\SAscenesmithN{31}
\def\SAmolmospacesN{636}

\def\SBversion{next-superunified-single-s39-0c15b742b912-1000-all}
\def\SBN{1{,}000}
\def\SBprocthorObj{73.4}
\def\SBinfinigenObj{143.9}
\def\SBholoqwenObj{87.3}
\def\SBhologemmaObj{93.2}
\def\SBoursObj{88.8}
\def\SBprocthorValid{33.4}
\def\SBinfinigenValid{73.9}
\def\SBholoqwenValid{76.5}
\def\SBhologemmaValid{82.0}
\def\SBoursValid{\textbf{99.1}}
\def\SBprocthorReach{96.1}
\def\SBinfinigenReach{98.8}
\def\SBholoqwenReach{89.7}
\def\SBhologemmaReach{93.1}
\def\SBoursReach{\textbf{100.0}}
\def\SBprocthorACC{98.7}
\def\SBinfinigenACC{90.8}
\def\SBholoqwenACC{97.7}
\def\SBhologemmaACC{97.8}
\def\SBoursACC{\textbf{99.6}}
\def\SBprocthorNAV{\textbf{99.8}}
\def\SBinfinigenNAV{96.3}
\def\SBholoqwenNAV{99.2}
\def\SBhologemmaNAV{99.2}
\def\SBoursNAV{99.7}
\def\SBprocthorCOL{3.69}
\def\SBinfinigenCOL{6.39}
\def\SBholoqwenCOL{2.97}
\def\SBhologemmaCOL{5.25}
\def\SBoursCOL{\textbf{0.36}}
\def\SBprocthorOOB{0.02}
\def\SBinfinigenOOB{1.51}
\def\SBholoqwenOOB{0.10}
\def\SBhologemmaOOB{0.14}
\def\SBoursOOB{\textbf{0.00}}
\def\SBprocthorN{1000}
\def\SBinfinigenN{1000}
\def\SBholoqwenN{826}
\def\SBhologemmaN{778}
\def\SBoursN{1000}
\def\TVlayprocthorRealWin{72.5}
\def\TVlayprocthorRealCI{[66, 78]}
\def\TVlayprocthorRealH{0.47}
\def\TVlayprocthorRealSig{yes}
\def\TVlayprocthorRealN{200}
\def\TVlayprocthorAdhWin{69.5}
\def\TVlayprocthorAdhCI{[63, 76]}
\def\TVlayprocthorAdhH{0.40}
\def\TVlayprocthorAdhSig{yes}
\def\TVlayprocthorAdhN{197}
\def\TVlayinfinigenRealWin{40.5}
\def\TVlayinfinigenRealCI{[34, 47]}
\def\TVlayinfinigenRealH{-0.19}
\def\TVlayinfinigenRealSig{yes}
\def\TVlayinfinigenRealN{200}
\def\TVlayinfinigenAdhWin{45.2}
\def\TVlayinfinigenAdhCI{[38, 52]}
\def\TVlayinfinigenAdhH{-0.10}
\def\TVlayinfinigenAdhSig{no}
\def\TVlayinfinigenAdhN{197}
\def\TVlayholoqwenRealWin{54.5}
\def\TVlayholoqwenRealCI{[48, 61]}
\def\TVlayholoqwenRealH{0.09}
\def\TVlayholoqwenRealSig{no}
\def\TVlayholoqwenRealN{200}
\def\TVlayholoqwenAdhWin{48.0}
\def\TVlayholoqwenAdhCI{[41, 55]}
\def\TVlayholoqwenAdhH{-0.04}
\def\TVlayholoqwenAdhSig{no}
\def\TVlayholoqwenAdhN{200}
\def\TVlayhologemmaRealWin{46.0}
\def\TVlayhologemmaRealCI{[39, 53]}
\def\TVlayhologemmaRealH{-0.08}
\def\TVlayhologemmaRealSig{no}
\def\TVlayhologemmaRealN{200}
\def\TVlayhologemmaAdhWin{48.5}
\def\TVlayhologemmaAdhCI{[42, 55]}
\def\TVlayhologemmaAdhH{-0.03}
\def\TVlayhologemmaAdhSig{no}
\def\TVlayhologemmaAdhN{194}
\def\TVlayhdRealWin{77.5}
\def\TVlayhdRealCI{[75, 80]}
\def\TVlayhdRealH{0.58}
\def\TVlayhdRealSig{yes}
\def\TVlayhdRealN{1562}
\def\TVlayhdAdhWin{\na}
\def\TVlayhdAdhCI{}
\def\TVlayhdAdhH{\na}
\def\TVlayhdAdhSig{\na}
\def\TVlayhdAdhN{0}
\def\TVlaywallplanRealWin{67.0}
\def\TVlaywallplanRealCI{[65, 69]}
\def\TVlaywallplanRealH{0.35}
\def\TVlaywallplanRealSig{yes}
\def\TVlaywallplanRealN{1562}
\def\TVlaywallplanAdhWin{\na}
\def\TVlaywallplanAdhCI{}
\def\TVlaywallplanAdhH{\na}
\def\TVlaywallplanAdhSig{\na}
\def\TVlaywallplanAdhN{0}
\def\TVlayrealRealWin{31.5}
\def\TVlayrealRealCI{[25, 38]}
\def\TVlayrealRealH{-0.38}
\def\TVlayrealRealSig{yes}
\def\TVlayrealRealN{200}
\def\TVlayrealAdhWin{\na}
\def\TVlayrealAdhCI{}
\def\TVlayrealAdhH{\na}
\def\TVlayrealAdhSig{\na}
\def\TVlayrealAdhN{0}
\def\TVsceprocthorRealWin{85.5}
\def\TVsceprocthorRealCI{[80, 90]}
\def\TVsceprocthorRealH{0.79}
\def\TVsceprocthorRealSig{yes}
\def\TVsceprocthorRealN{200}
\def\TVsceprocthorAdhWin{83.5}
\def\TVsceprocthorAdhCI{[78, 88]}
\def\TVsceprocthorAdhH{0.73}
\def\TVsceprocthorAdhSig{yes}
\def\TVsceprocthorAdhN{200}
\def\TVsceinfinigenRealWin{35.0}
\def\TVsceinfinigenRealCI{[29, 42]}
\def\TVsceinfinigenRealH{-0.30}
\def\TVsceinfinigenRealSig{yes}
\def\TVsceinfinigenRealN{200}
\def\TVsceinfinigenAdhWin{48.0}
\def\TVsceinfinigenAdhCI{[41, 55]}
\def\TVsceinfinigenAdhH{-0.04}
\def\TVsceinfinigenAdhSig{no}
\def\TVsceinfinigenAdhN{200}
\def\TVsceholoqwenRealWin{49.5}
\def\TVsceholoqwenRealCI{[43, 56]}
\def\TVsceholoqwenRealH{-0.01}
\def\TVsceholoqwenRealSig{no}
\def\TVsceholoqwenRealN{200}
\def\TVsceholoqwenAdhWin{65.0}
\def\TVsceholoqwenAdhCI{[58, 71]}
\def\TVsceholoqwenAdhH{0.30}
\def\TVsceholoqwenAdhSig{yes}
\def\TVsceholoqwenAdhN{200}
\def\TVscehologemmaRealWin{35.4}
\def\TVscehologemmaRealCI{[29, 42]}
\def\TVscehologemmaRealH{-0.30}
\def\TVscehologemmaRealSig{yes}
\def\TVscehologemmaRealN{198}
\def\TVscehologemmaAdhWin{57.1}
\def\TVscehologemmaAdhCI{[50, 64]}
\def\TVscehologemmaAdhH{0.14}
\def\TVscehologemmaAdhSig{no}
\def\TVscehologemmaAdhN{198}
\providecommand{\na}{\textcolor{gray}{n/a}}
\providecommand{\up}{$\uparrow$}
\providecommand{\dn}{$\downarrow$}


\begin{table*}[!htbp]
  \centering
  \caption{\textbf{\apm{} preserves the requested door graph in every RPLAN layout it returns.}
    Block~(a) reports results for 1{,}000 RPLAN apartments, giving each method the input it was designed to use.
Blocks~(b--d) group the 1{,}000 single-story homes of \cref{tab:scenes} by the style drawn by our generator (\RSN{} Swiss, \RFN{} Finnish, and \RUN{} US).
Each group is compared with real layouts of the same style, and methods without brief input receive our architectural program.
    The \emph{Real} row evaluates real room layouts at the same sample size.
Both distribution metrics subtract this reference score, giving $\Delta$FID and $\Delta$KID ($\times10^3$).
    \Cref{sec:layout-benchmark} defines the inputs, reference data, and metrics.
An entry is \na{} when the method cannot use the input or the metric cannot be measured.
Per column and block, among the methods: \rankkey{rankfirst}{\textbf{best}}, \rankkey{ranksecond}{second} and \rankkey{rankthird}{third best}.}
  \label{tab:layouts}
  \small
  \setlength{\tabcolsep}{4pt}
  \resizebox{\textwidth}{!}{%
  \begin{tabular}{llr|cc|ccc|cc}
    \toprule
    & & & \multicolumn{2}{c|}{Floor plan} & \multicolumn{3}{c|}{Input adherence} & \multicolumn{2}{c}{Realism} \\
    Method & Input & $N$ & Valid\,\%\up & Reach\,\%\up & Types\,\%\up & Graph\,\%\up & Area\dn & $\Delta$FID\dn & $\Delta$KID\dn \\
    \midrule
    \multicolumn{10}{l}{\emph{(a) RPLAN, 1{,}000 apartments (Graph2Plan test split; reference: RPLAN). Valid: \% of returned layouts / \% of all inputs.}} \\
    Real (RPLAN) & -- & \RArealN & \RArealValid{} / -- & \RArealReach & \RArealTypes & \RArealGraph & \RArealArea & \RArealDFID & \RArealDKID \\
    House-GAN++~\cite{nauata2021house} & bubble & \RAhgppN & \RAhgppValid{} / 44.6 & \RAhgppReach & \RAhgppTypes & \RAhgppGraph & \RAhgppArea & \RAhgppDFID & \RAhgppDKID \\
    HouseDiffusion~\cite{shabani2023housediffusion} & bubble & \RAhdN & \RAhdValid{} / 60.1 & \RAhdReach & \RAhdTypes & \RAhdGraph & \RAhdArea & \RAhdDFID & \RAhdDKID \\
    WallPlan~\cite{sun2022wallplan} & boundary & \RAwallplanN & \RAwallplanValid{} / 87.0 & \RAwallplanReach & \RAwallplanTypes & \RAwallplanGraph & \RAwallplanArea & \RAwallplanDFID & \RAwallplanDKID \\
    ProcTHOR~\cite{deitke2022procthor} & bubble+areas & \RAprocthorN & \RAprocthorValid{} / 36.8 & \RAprocthorReach & \RAprocthorTypes & \RAprocthorGraph & \RAprocthorArea & \RAprocthorDFID & \RAprocthorDKID \\
    Infinigen Indoors~\cite{raistrick2024infinigen} & bubble+areas & \RAinfinigenN & \RAinfinigenValid{} / 76.8 & \RAinfinigenReach & \RAinfinigenTypes & \RAinfinigenGraph & \RAinfinigenArea & \RAinfinigenDFID & \RAinfinigenDKID \\
    Holodeck~\cite{yang2024holodeck} (Qwen3.8-27B) & bubble+areas & \RAholoqwenN & \RAholoqwenValid{} / 61.4 & \RAholoqwenReach & \RAholoqwenTypes & \RAholoqwenGraph & \RAholoqwenArea & \RAholoqwenDFID & \RAholoqwenDKID \\
    Holodeck~\cite{yang2024holodeck} (Gemma-4-31B) & bubble+areas & \RAhologemmaN & \RAhologemmaValid{} / 33.7 & \RAhologemmaReach & \RAhologemmaTypes & \RAhologemmaGraph & \RAhologemmaArea & \RAhologemmaDFID & \RAhologemmaDKID \\
    \apm (ours) & bubble & \RAapmbubbleN & \RAapmbubbleValid{} / 77.3 & \RAapmbubbleReach & \RAapmbubbleTypes & \RAapmbubbleGraph & \RAapmbubbleArea & \RAapmbubbleDFID & \RAapmbubbleDKID \\
    \apm (ours) & bubble+areas & \RAapmareasN & \RAapmareasValid{} / 73.2 & \RAapmareasReach & \RAapmareasTypes & \RAapmareasGraph & \RAapmareasArea & \RAapmareasDFID & \RAapmareasDKID \\
    \midrule
    \multicolumn{10}{l}{\emph{(b) Brief, single-story homes in Swiss Dwellings style, \RSN{} briefs (reference: composition-matched Swiss Dwellings)}} \\
    Real & -- & \RSoursN & \RSrealValid & \na & \na & \na & \na & 0 & 0 \\
    ProcTHOR~\cite{deitke2022procthor} & program & \RSprocthorN & \RSprocthorValid & \RSprocthorReach & \RSprocthorTypes & \RSprocthorGraph & \RSprocthorArea & \RSprocthorDFID & \RSprocthorDKID \\
    Infinigen Indoors~\cite{raistrick2024infinigen} & program & \RSinfinigenN & \RSinfinigenValid & \RSinfinigenReach & \RSinfinigenTypes & \RSinfinigenGraph & \RSinfinigenArea & \RSinfinigenDFID & \RSinfinigenDKID \\
    Holodeck~\cite{yang2024holodeck} (Qwen3.8-27B) & brief (text) & \RSholoqwenN & \RSholoqwenValid & \RSholoqwenReach & \RSholoqwenTypes & \RSholoqwenGraph & \RSholoqwenArea & \RSholoqwenDFID & \RSholoqwenDKID \\
    Holodeck~\cite{yang2024holodeck} (Gemma-4-31B) & brief (text) & \RShologemmaN & \RShologemmaValid & \RShologemmaReach & \RShologemmaTypes & \RShologemmaGraph & \RShologemmaArea & \RShologemmaDFID & \RShologemmaDKID \\
    \apm (ours) & brief & \RSoursN & \RSoursValid & \RSoursReach & \RSoursTypes & \RSoursGraph & \RSoursArea & \RSoursDFID & \RSoursDKID \\
    \midrule
    \multicolumn{10}{l}{\emph{(c) Brief, single-story homes in Finnish style, \RFN{} briefs (reference: composition-matched CubiCasa5k)}} \\
    Real & -- & \RFoursN & \RFrealValid & \na & \na & \na & \na & 0 & 0 \\
    ProcTHOR~\cite{deitke2022procthor} & program & \RFprocthorN & \RFprocthorValid & \RFprocthorReach & \RFprocthorTypes & \RFprocthorGraph & \RFprocthorArea & \RFprocthorDFID & \RFprocthorDKID \\
    Infinigen Indoors~\cite{raistrick2024infinigen} & program & \RFinfinigenN & \RFinfinigenValid & \RFinfinigenReach & \RFinfinigenTypes & \RFinfinigenGraph & \RFinfinigenArea & \RFinfinigenDFID & \RFinfinigenDKID \\
    Holodeck~\cite{yang2024holodeck} (Qwen3.8-27B) & brief (text) & \RFholoqwenN & \RFholoqwenValid & \RFholoqwenReach & \RFholoqwenTypes & \RFholoqwenGraph & \RFholoqwenArea & \RFholoqwenDFID & \RFholoqwenDKID \\
    Holodeck~\cite{yang2024holodeck} (Gemma-4-31B) & brief (text) & \RFhologemmaN & \RFhologemmaValid & \RFhologemmaReach & \RFhologemmaTypes & \RFhologemmaGraph & \RFhologemmaArea & \RFhologemmaDFID & \RFhologemmaDKID \\
    \apm (ours) & brief & \RFoursN & \RFoursValid & \RFoursReach & \RFoursTypes & \RFoursGraph & \RFoursArea & \RFoursDFID & \RFoursDKID \\
    \midrule
    \multicolumn{10}{l}{\emph{(d) Brief, single-story homes in US style, \RUN{} briefs (reference: composition-matched ZInD)}} \\
    Real & -- & \RUoursN & \RUrealValid & \na & \na & \na & \na & 0 & 0 \\
    ProcTHOR~\cite{deitke2022procthor} & program & \RUprocthorN & \RUprocthorValid & \RUprocthorReach & \RUprocthorTypes & \RUprocthorGraph & \RUprocthorArea & \RUprocthorDFID & \RUprocthorDKID \\
    Infinigen Indoors~\cite{raistrick2024infinigen} & program & \RUinfinigenN & \RUinfinigenValid & \RUinfinigenReach & \RUinfinigenTypes & \RUinfinigenGraph & \RUinfinigenArea & \RUinfinigenDFID & \RUinfinigenDKID \\
    Holodeck~\cite{yang2024holodeck} (Qwen3.8-27B) & brief (text) & \RUholoqwenN & \RUholoqwenValid & \RUholoqwenReach & \RUholoqwenTypes & \RUholoqwenGraph & \RUholoqwenArea & \RUholoqwenDFID & \RUholoqwenDKID \\
    Holodeck~\cite{yang2024holodeck} (Gemma-4-31B) & brief (text) & \RUhologemmaN & \RUhologemmaValid & \RUhologemmaReach & \RUhologemmaTypes & \RUhologemmaGraph & \RUhologemmaArea & \RUhologemmaDFID & \RUhologemmaDKID \\
    \apm (ours) & brief & \RUoursN & \RUoursValid & \RUoursReach & \RUoursTypes & \RUoursGraph & \RUoursArea & \RUoursDFID & \RUoursDKID \\
    \bottomrule
  \end{tabular}}
\end{table*}

\begin{figure*}[!htbp]
  \centering
  \includegraphics[width=\linewidth]{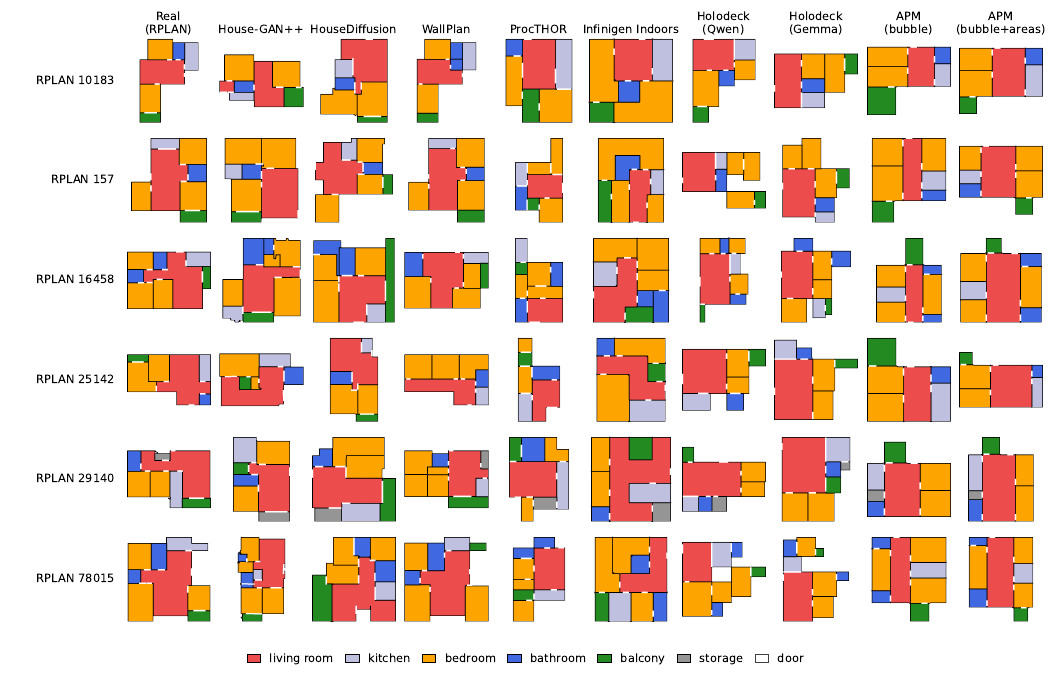}
  \caption{\textbf{Room layouts for RPLAN inputs} (block~(a) of \cref{tab:layouts}). Each row is one of the 1{,}000 RPLAN test inputs; the first column is its real layout. All plans are drawn the same way, whatever the method's export convention: rooms are colored by type, gaps of up to 0.6\,m between rooms (walls in exports with net room areas) are closed, walls have one thickness, and each door is drawn as a white opening of its width on the nearest wall. Plans are scaled to fit their cell, since House-GAN++ and HouseDiffusion have no metric scale.} 
  \label{fig:layouts_rplan}
\end{figure*}

\begin{figure*}[!htbp]
  \centering
  \includegraphics[width=\linewidth]{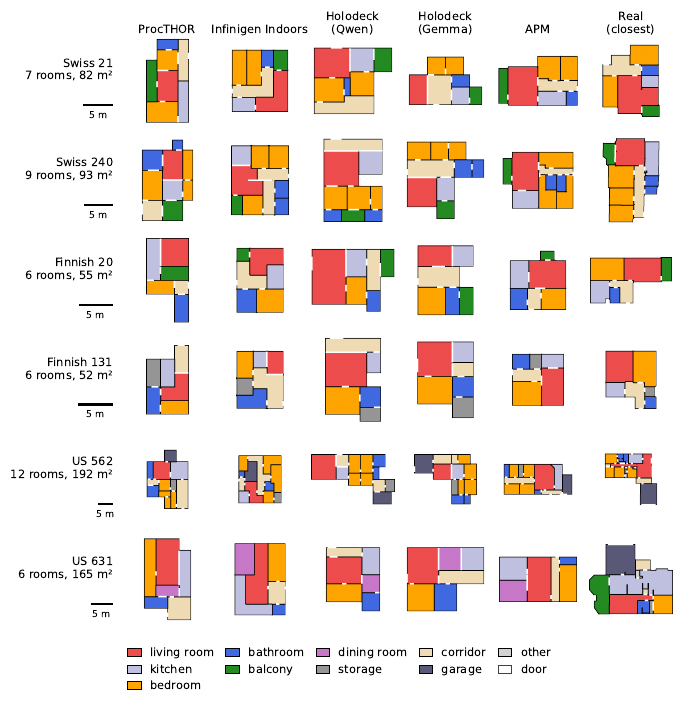}
  \caption{\textbf{Room layouts for briefs} (blocks~(b--d) of \cref{tab:layouts}): two briefs per style from the 1{,}000 single-story homes, each with its room count and net area. \apm{} generated these briefs, so no real layout of the same brief exists; the last column shows the layout from the style's reference dataset (Swiss Dwellings, CubiCasa5k or ZInD) that is closest to the \apm{} home in room types and total area. Plans are drawn as in \cref{fig:layouts_rplan}; all plans in a row share one metric scale (5\,m bar).}
  \label{fig:layouts_briefs}
\end{figure*}

\subsection{Layout validity and constraint failures}
We now return to the complete, furnished homes evaluated in \ifdefined\includeapmsupplement\cref{tab:scenes}\else Table~1 of the main paper\fi{}, block~(b) to examine which floor-plan checks fail and whether furniture blocks access through doors.
The G-program can reject a floor plan during construction (\cref{sec:supp_dev_yield}), so the validity rate in the table counts only completed homes.
\Cref{tab:supp_valid_breakdown} shows how often each method's completed homes fail each floor-plan validity check defined in \cref{sec:supp_scene_metrics}.
Nine of the 1{,}000 completed \apm{} homes fail, all on room size.
\Cref{tab:supp_valid_rooms} breaks down room-size failures by room type: five \apm{} living rooms fall below the minimum width and four bathrooms fall below the minimum area.
Even when a floor plan connects two rooms through a doorway, furniture can block the passage.
The D-program flags blocked doors in 0.3\% of \apm{} homes, compared with 24.0\% for ProcTHOR and 55.2\% for Infinigen Indoors (\cref{tab:supp_constraint_fails}).
\Cref{sec:supp_constraints,sec:supp_scene_metrics} define the checks and explain how we handle missing information.
\begin{table}[!htbp]
  \centering
  \small
  \setlength{\tabcolsep}{1.8pt}
  \caption{\textbf{Narrow rooms and small rooms account for all nine invalid \apm{} layouts.}
The first column gives the percentage of invalid homes among the completed homes in \ifdefined\includeapmsupplement\cref{tab:scenes}\else Table~1 of the main paper\fi.
The remaining columns show which criteria each method fails, using the thresholds in \cref{tab:supp_valid}.
A home can fail more than one criterion.}
  \label{tab:supp_valid_breakdown}
  \begin{tabular}{lrrrrrrrr}
    \toprule
    & & & \multicolumn{3}{c}{Layout} & \multicolumn{3}{c}{Room size} \\
    \cmidrule(lr){4-6}\cmidrule(lr){7-9}
    Method & $N$ & Invalid & Overlap & Holes & Discon- & Aspect & Area & Width \\
    & & & $>$2\% & $>$2\% & nected & & & \\
    \midrule
    ProcTHOR & 1{,}000 & 66.6 & 0.0 & 0.0 & 0.0 & 29.8 & 11.8 & 57.2 \\
    Infinigen Indoors & 1{,}000 & 26.1 & 0.0 & 0.0 & 0.0 & 11.8 & 6.8 & 16.5 \\
    Holodeck (Qwen) & 826 & 23.5 & 8.1 & 14.4 & 0.2 & 2.1 & 0.4 & 0.6 \\
    Holodeck (Gemma) & 778 & 18.0 & 15.6 & 0.8 & 0.0 & 1.8 & 0.3 & 0.4 \\
    SceneSmith & 31 & 9.7 & 0.0 & 9.7 & 0.0 & 0.0 & 0.0 & 0.0 \\
    MolmoSpaces & 636 & 15.6 & 2.0 & 9.6 & 0.0 & 0.6 & 1.4 & 3.6 \\
    \midrule
    \apm (ours) & 1{,}000 & 0.9 & 0.0 & 0.0 & 0.0 & 0.0 & 0.4 & 0.5 \\
    \bottomrule
  \end{tabular}
\end{table}

\begin{table}[!htbp]
  \centering
  \small
  \setlength{\tabcolsep}{2.5pt}
  \caption{\textbf{Room-size failures depend on the room type.}
Of the nine \apm{} rooms below the thresholds in \cref{tab:supp_valid}, five are too-narrow living rooms and four are undersized bathrooms.
For each method, we report the total number of failing rooms and break down the most frequently failing types by threshold.
A room can fail more than one threshold.
Layout-level overlaps and holes appear in \cref{tab:supp_valid_breakdown}.}
  \label{tab:supp_valid_rooms}
  \begin{tabular}{lrp{0.56\columnwidth}}
    \toprule
    Method & Rooms & Most frequent failing room types \\
    \midrule
    ProcTHOR & 997/7{,}593 & \raggedright bedroom 620/1{,}753 (width 501, aspect 265, area 58); living room 261/1{,}000 (width 257, area 59, aspect 12)\tabularnewline
    Infinigen Indoors & 312/7{,}593 & \raggedright bedroom 147/1{,}753 (aspect 101, width 46, area 29); living room 106/1{,}000 (width 105, area 21, aspect 3)\tabularnewline
    Holodeck (Qwen) & 23/6{,}018 & \raggedright bathroom 13/1{,}073 (aspect 12, area 1); bedroom 4/1{,}352 (width 3, aspect 2, area 1)\tabularnewline
    Holodeck (Gemma) & 15/5{,}658 & \raggedright bathroom 12/1{,}001 (aspect 12); bedroom 2/1{,}248 (width 2, aspect 2, area 1)\tabularnewline
    SceneSmith & 0/185 & \raggedright --\tabularnewline
    MolmoSpaces & 30/4{,}531 & \raggedright bedroom 15/1{,}249 (width 13, area 7, aspect 1); living room 7/608 (width 7, area 3)\tabularnewline
    \midrule
    \apm (ours) & 9/7{,}593 & \raggedright living room 5/1{,}000 (width 5); bathroom 4/1{,}345 (area 4)\tabularnewline
    \bottomrule
  \end{tabular}
\end{table}

\begin{table*}[!htbp]
  \centering
  \small
  \setlength{\tabcolsep}{4.5pt}
  \caption{\textbf{Blocked doors and missing fixtures distinguish the methods.}
Each row gives the percentage of graded homes failing a check, with the number of unknown outcomes in parentheses.
We grade 1{,}000 homes each for \apm{}, ProcTHOR, and Infinigen, 758 of 826 for Holodeck with Qwen3.8-27B, 649 of 778 with Gemma-4-31B, all 31 for SceneSmith, and 613 of 636 for MolmoSpaces.
The remaining homes could not be represented in our evaluation format.}
  \label{tab:supp_constraint_fails}
  \begin{tabular}{lrrrrrrr}
    \toprule
    Check & \apm & ProcTHOR & Infinigen & Holo.\,Q & Holo.\,G & SceneSmith & Molmo. \\
    \midrule
    \texttt{all-rooms-reachable} & 0.0 & 3.9 & 0.0 & 8.3 & 2.0 & 3.2 & 0.0 \\
    \texttt{room-has-entrance} & 0.0 & 0.0 & 0.0 & 0.4\,(1) & 0.0\,(3) & 3.2 & 0.0 \\
    \texttt{doors-not-blocked} & 0.3 & 24.0 & 55.2\,(40) & 4.2\,(4) & 4.9\,(4) & 29.0\,(1) & 2.0 \\
    \texttt{openings-clear} & 0.3 & 24.0 & 55.2\,(40) & 4.2\,(4) & 4.9\,(4) & 29.0\,(1) & 2.0 \\
    \texttt{bathroom-has-toilet-or-sink} & 0.0 & 97.4 & 88.6 & 71.8 & 75.3 & 58.1 & 68.0 \\
    \texttt{home-has-toilet} & 0.0 & 7.8 & 0.7 & 18.5 & 12.5 & 9.7 & 85.0 \\
    \texttt{kitchen-has-sink} & 0.0 & 100.0 & 80.7 & 82.1 & 94.1 & 12.9 & 73.2 \\
    \texttt{bedroom-has-bed} & 0.0 & 48.1 & 39.9 & 9.0 & 0.6 & 0.0 & 43.7 \\
    \texttt{required-room-furniture} & 0.0 & 100.0 & 100.0 & 99.9 & 100.0 & 74.2 & 98.2 \\
    \texttt{furniture-inside-floor} & 0.0 & 25.8 & 55.3 & 26.0 & 41.6 & 0.0 & 28.4 \\
    \texttt{furniture-no-overlap} & 0.0 & 85.0 & 87.5 & 74.8 & 87.4 & 100.0 & 88.7 \\
    \texttt{wall-anchors-touch} & 0.0 & 0.0 & 0.0 & 0.0 & 0.0 & 0.0 & 0.0 \\
    \texttt{hanging-attachments-supported} & 0.0 & 0.0 & 0.0 & 0.0 & 0.0 & 0.0 & 0.0 \\
    \texttt{decoration-supported} & 0.0 & 0.0\,(1000) & 0.0\,(1000) & 0.0\,(755) & 0.0\,(647) & 0.0\,(31) & 0.0\,(612) \\
    \texttt{bathroom-sink-rim-supported} & 0.0 & 0.0\,(854) & 0.0\,(987) & 0.0\,(601) & 0.0\,(437) & 0.0\,(28) & 0.0\,(101) \\
    \texttt{bed-headboard-against-wall} & 0.0 & 24.2\,(203) & 55.6\,(3) & 61.1\,(101) & 70.1\,(43) & 35.5 & 49.4\,(59) \\
    \texttt{seats-face-into-room} & 0.0 & 13.3 & 9.9 & 19.7 & 8.0 & 19.4 & 8.0 \\
    \texttt{window-coverings-fit} & 0.0 & 0.0 & 0.0 & 0.0 & 0.0 & 0.0 & 0.0 \\
    \texttt{storage-front-access} & 0.1 & 33.7 & 98.3 & 71.4 & 67.8 & 74.2 & 69.2 \\
    \texttt{closet-layout-shape} & 0.1 & 23.8 & 20.8 & 10.3 & 14.6\,(1) & 0.0 & 2.0 \\
    \texttt{closet-storage-occupancy} & 0.9 & 23.8 & 23.5 & 21.0 & 23.6 & 3.2 & 2.6 \\
    \texttt{room-program-plausibility} & 10.7\,(100) & 72.5\,(12) & 47.4\,(37) & 39.2\,(27) & 34.4\,(7) & 6.5\,(19) & 18.4\,(360) \\
    \texttt{no-adjacent-bathrooms} & 0.0 & 6.3 & 0.0 & 7.9 & 8.6 & 0.0 & 8.8 \\
    \texttt{wet-room-floor-not-wood} & 0.0 & 0.0\,(1000) & 0.0\,(1000) & 0.0\,(758) & 0.0\,(649) & 0.0\,(28) & 0.0\,(447) \\
    \bottomrule
  \end{tabular}
\end{table*}
\subsection{Perceptual Study}
\label{sec:supp_perceptual}
We assessed the realism of the furnished homes by comparing human judgments with those of two vision-language models on the same image pairs.

\subsubsection{User Study}
Participants saw two renderings side by side, one of a home generated by \apm{} and one by a baseline for the same design brief, and they were asked to judge which looked more like a real, lived-in home.
Against each of the four baselines, participants chose the \apm{} home more often.

\paragraph{Pairs.}
For each baseline, \apm's home is paired with the baseline's home for 100 briefs of \ifdefined\includeapmsupplement\cref{tab:scenes}\else Table~1\fi\,(b) for which every method produced a scene.
Each home is shown as one perspective top-down view rendered by one Blender pipeline with the same walls, materials, lighting and camera for every method.
Left and right placement is balanced: \apm is on the left in 50 of the 100 pairs.
The question is \emph{``Which home looks more like a real, lived-in home?''}, with the answers A, B or equal (\cref{fig:study_page}).

\begin{figure}[!htbp]
  \centering
  \includegraphics[width=\linewidth]{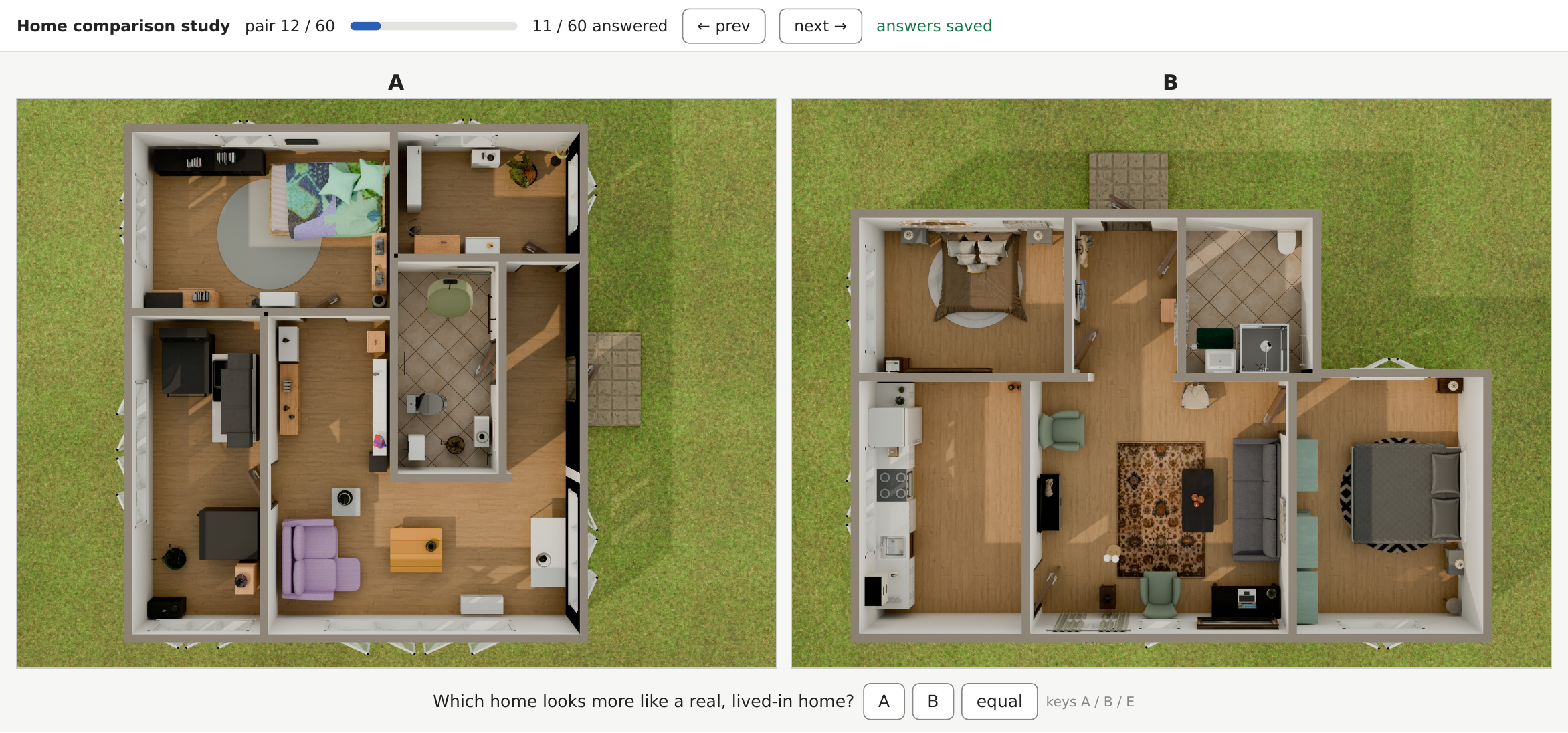}
  \caption{\textbf{The study page.} One pair as participants saw it, here Infinigen Indoors (A) and \apm (B) for the same brief. Participants saw only the letters A and B.}
  \label{fig:study_page}
\end{figure}


\paragraph{Participants and Responses}
We recruited volunteers for an anonymous web study by circulating an open invitation.
We excluded three of the 38 respondents because their median response time was below 1.5\,s per pair.
Among the remaining 35, 13 reported professional experience in architecture or interior design and 10 in 3D modelling or computer graphics.
Participants answer the same question on the same pairs in an online questionnaire; each participant is assigned 60 image pairs in random order, with 15 pairs comparing \apm{} to each baseline (ProcTHOR, Infinigen Indoors, Holodeck with Qwen, and Holodeck with Gemma).
We included responses from participants who stopped before completing all 60 pairs, so participants contributed different numbers of judgements and the totals varied by baseline.
The 35 participants gave 1{,}053 judgements (257--272 per comparison); 15 of them finished all 60 pairs.
Before the first pair, participants read a consent page (purpose, duration, anonymity, voluntary participation with the option to stop at any time) and agreed to take part; we stored no names, e-mail addresses or other identifying data.

\paragraph{User Study Results}
Participants preferred \apm in 67.4\% of all judgements and the baseline in 17.2\%, and answered ``equal'' in 15.4\%.
Excluding ``equal'', the share preferring \apm is 93.3\% against ProcTHOR, 70.4\% against Infinigen Indoors, 84.4\% against Holodeck with Qwen3.8-27B and 70.5\% against Holodeck with Gemma-4-31B (\ifdefined\includeapmsupplement\cref{tab:userstudy}\else Table~2 of the main paper\fi), with 95\% Wilson confidence intervals.
Preference does not depend on self-reported expertise: excluding ``equal'', participants with professional experience in architecture or interior design (13 of 35) preferred \apm{} in 79.9\% of their judgements, those with professional experience in 3D modelling or computer graphics (10) in 77.7\%, and the others in 76--87\%.

\subsubsection{VLM Judges}
Each of the 35 volunteers judged a subset of the image pairs.
We asked two vision-language models to judge all 100 pairs per baseline using the same question to see whether their overall preferences matched those of the volunteers.
Claude Sonnet~5.5 and GPT-6.1 Sol judge every pair once, in the study's order, through their batch interfaces with high reasoning effort and images of 1{,}568\,px.
We report each model's share of judgements favoring \apm{} and its 95\% Wilson confidence interval (\ifdefined\includeapmsupplement\cref{tab:userstudy}\else Table~2 of the main paper\fi).
Sonnet~5.5 preferred \apm in 100.0\% of the pairs against ProcTHOR, 83.0\% against Infinigen Indoors, 75.0\% against Holodeck with Qwen3.8-27B and 73.0\% against Holodeck with Gemma-4-31B; GPT-6.1 Sol in 97.0\%, 86.0\%, 76.0\% and 67.0\%.
Both models favored \apm{} over all four baselines, as did the participants, but neither model chose ``equal'', an answer participants gave in 15.4\% of their judgements.

\section{Generator development}
This section describes how we judged whether the G-program improved. We first trace how many homes passed all hard constraints across successive versions (\cref{sec:supp_dev_yield}), then show how missed defects, mismatched reference measurements and user preferences led us to revise the scoring objectives (\cref{sec:supp_objective_trace}).

\label{sec:supp_development}
\subsection{Completed homes in the development runs}
\label{sec:supp_dev_yield}
Stricter checks reduced the share of attempts that produce a home, but the revised G-program delivered more homes that pass all checks.

Each version of the G-program and each ablation is run on the same 300 development seeds.
An attempt can fail to produce a home, for example when a layout gate rejects the floor plan, so the shares in \ifdefined\includeapmsupplement\cref{fig:trajectory} and \cref{tab:ablation}\else Fig.~5 and Table~3 of the main paper\fi{} are computed over the completed homes.
\Cref{tab:supp_dev_yield} gives the completed homes and those passing all hard constraints.
Version s18 was not run on these seeds.
Fewer homes complete in the later versions (69--75\% from s04 on, against 81\% at s00) because the added layout checks reject floor plans; counted per 300 attempts, s10 still delivers 176 passing homes, against 83 without D-program feedback.
\begin{table}[!htbp]
  \centering
  \scriptsize
  \setlength{\tabcolsep}{3pt}
  \caption{\textbf{More homes pass all constraints as the generator improves.}
Each G-program version was run on the same 300 seeds.
The table counts completed homes and, among those, homes passing every hard constraint.}
  \label{tab:supp_dev_yield}

  \begin{tabular}{lrrr@{\hspace{10pt}}lrrr}
    \toprule
    Version & Compl. & Pass & \% & Version & Compl. & Pass & \% \\
    \midrule
    s00 & 242 & 62 & 25.6 & s10 & 208 & 176 & 84.6 \\
    s01 & 231 & 64 & 27.7 & s11 & 215 & 201 & 93.5 \\
    s02 & 225 & 70 & 31.1 & s12 & 215 & 205 & 95.3 \\
    s03 & 214 & 74 & 34.6 & s13 & 218 & 208 & 95.4 \\
    s04 & 215 & 119 & 55.3 & s14 & 224 & 214 & 95.5 \\
    s05 & 215 & 142 & 66.0 & s15 & 224 & 216 & 96.4 \\
    s06 & 215 & 152 & 70.7 & s16 & 223 & 211 & 94.6 \\
    s07 & 215 & 150 & 69.8 & s17 & 223 & 219 & 98.2 \\
    s08 & 210 & 154 & 73.3 & s19 & 224 & 220 & 98.2 \\
    s09 & 211 & 178 & 84.4 & & & & \\
    \bottomrule
  \end{tabular}
\end{table}
\setcounter{apmsavedtable}{\value{table}}\setcounter{table}{0}
\subsection{Recorded revisions to the scoring objectives} 
\label{sec:supp_objective_trace}
A scoring function can reward the wrong behavior even when its measurements are computed correctly.
Apartments and US homes are scored separately, using reference data and active measurements appropriate to each population.
\Cref{tab:objective_trace_apartments,tab:objective_trace_us} show how visual inspection, reference-data audits and user guidance changed the scoring objectives for apartments and US homes.

Rows follow the first recorded change, with related later changes grouped together.
These selected events changed score components or their treatment in scoring; none changed the hard constraints required in those runs. 


\begin{table*}[!htbp]
\centering
\small
\setlength{\tabcolsep}{5pt}
\renewcommand{\arraystretch}{1.14}
\caption{\textbf{Observed defects and mismatches changed what the apartment objective measured.}
Each row connects an observed problem to the change made to assess apartment G-programs more appropriately.
The sequence includes changes motivated by the user's preference, inconsistent reference measurements, and defects found during visual inspection.}
\label{tab:objective_trace_apartments}
\begin{tabular}{@{}p{0.46\textwidth}p{0.50\textwidth}@{}}
\toprule
Observation and reason for revision & Change to the apartment objective \\
\midrule
A candidate replaced styled items with neutral ones.
Too few scenes then qualified for a style comparison, which became inactive and raised the score.
Losing measurement coverage had been rewarded.
&
\textbf{New measurement and coverage treatment.}
Added the share of styled items.
A comparison that loses substantial coverage receives the maximum discrepancy rather than disappearing.
A later refinement distinguished a real coverage loss from a small change across the sample-count threshold.
\\ \addlinespace
The user requested strongly furnished homes, but the score penalized furniture density above the reference mean.
After density was corrected, object-presence comparisons still penalized the additional furniture.
&
\textbf{Preference-aligned targets.}
Allowed densities from the reference value up to its 90th percentile without penalty, first in living rooms and bedrooms and then in bathrooms.
Object-presence targets for living rooms and bedrooms were subsequently based on reference rooms within the accepted density band.
\\ \addlinespace
Generated bathrooms counted all cabinets, whereas the HSSD reference counted only one cabinet category.
This made ordinary bathroom storage appear excessive.
&
\textbf{Corrected reference measurement.}
Recomputed reference cabinet presence using the same categories as the generated scenes, so the comparison measured the same property on both sides.
\\ \addlinespace
Visual inspection found the same painting repeated within a room.
The style-related score had not captured this defect.
&
\textbf{New measurement.}
Added the fraction of rooms with repeated paintings, compared with HSSD.
The candidate that exposed the problem was rejected during visual review.
\\ \addlinespace
Visual review found four identical wardrobes in one bedroom.
As with repeated paintings, the existing comparisons missed an implausible arrangement.
&
\textbf{New measurement.}
Added the fraction of bedrooms with at least three wardrobes, compared with 3D-FRONT.
This supplied feedback for revising wardrobe selection in subsequent candidates.
\\
\bottomrule
\end{tabular}
\end{table*}

\begin{table*}[!htbp]
\centering
\small
\setlength{\tabcolsep}{5pt}
\renewcommand{\arraystretch}{1.14}
\caption{\textbf{Further recorded revisions to the US-home objective.}
The US-home objective also received the coverage treatment and the bathroom-cabinet correction of \cref{tab:objective_trace_apartments}, and its furniture-density targets were relaxed in the same way; object-presence targets stayed unchanged because too few reference rooms qualified.
The rows list the remaining revisions, both based on ZInD.}
\label{tab:objective_trace_us}
\begin{tabular}{@{}p{0.46\textwidth}p{0.50\textwidth}@{}}
\toprule
Observation and reason for revision & Change to the US-home objective \\
\midrule
Generated bedrooms appeared too small, but an audit found that some ZInD bedroom polygons also contained closets, bathrooms, or halls.
The generator represented these as separate rooms.
&
\textbf{Corrected reference subset.}
Restricted bedroom-area comparisons to reference bedrooms without merged spaces, matching the generator's room definition.
\\ \addlinespace
Review of a generated US home found hallways and corridors taking 56\% of the indoor floor area.
No comparison measured this: 18\% of the generated US homes devoted more than 30\% of their indoor area to circulation, against 1.3\% of ZInD homes.
&
\textbf{New measurement.}
Added the share of homes whose hallways and corridors exceed 30\% of the indoor area, close to the 99th percentile of ZInD (31.5\%).
\\
\bottomrule
\end{tabular}
\end{table*}

The wardrobe case shows how visual review reveals a limitation of the numerical assessment.
One candidate met the numerical acceptance criteria, but review of selected renderings found four identical wardrobes in one bedroom.
The Orchestrator judged the arrangement excessive, rejected the candidate, and approved the new measurement in the last row of \cref{tab:objective_trace_apartments}.
The next candidate sampled wardrobe counts from the 3D-FRONT distribution, passed the acceptance procedure under the revised score, and was accepted.

\section{Evaluation protocol}
\label{sec:supp_eval}
\subsection{Baseline configurations}
\label{sec:supp_baselines}
\paragraph{ProcTHOR.}
Its room-layout stage builds the floor plan from our program; its furnishing stages then place AI2-THOR assets.
ProcTHOR furnishes only kitchens, living rooms, bedrooms and bathrooms, and its asset set has no stove or oven.
\paragraph{Infinigen Indoors.}
We run version 1.19 with its reduced object-solver setting \texttt{fast\_solve} (100/40/5 instead of 300/200/50 solver steps); default steps take about 10\,h per home, which is not feasible for 1{,}000 homes.
The floor-plan search keeps its defaults but is restricted to our program (\emph{strict}): the program's room graph, a fixed outline sized from the target areas, and a per-room area term.
Each home gets up to three attempts with new seeds; all 1{,}000 homes were generated.
\paragraph{Holodeck.}
Qwen3.8-27B and Gemma-4-31B are served locally with Holodeck's prompts and settings unchanged.
The input is the brief as text: the number of rooms of each type and the total net area.
Holodeck makes up to three attempts per brief when its own check rejects a floor plan.
Its small-object stage spawns repeated assets under one identifier, which stacks copies at one spot; we give each placement a unique identifier.
With Qwen3.8-27B, 172 of the 1{,}000 briefs got no floor plan: Holodeck's check rejected all three attempts, mostly for overlapping rooms, a room corner inside another room, angles below 90$^\circ$ or a room that shares no wall with the others (a few replies could not be parsed); 2 more failed while placing objects, which leaves 826 homes.
With Gemma-4-31B, 222 briefs gave no home within three attempts, which leaves 778; 182 of them succeed with up to ten attempts, but we keep the same limit for both models.
Failures grow with the size of the brief: the failed briefs have 9.1 (Qwen) and 8.7 (Gemma) rooms on average, the delivered ones 7.3; briefs with at most five rooms almost always succeed (4 and 1 failures of 141), while half and 38\% of those with twelve or more rooms fail.
Holodeck's rows in \ifdefined\includeapmsupplement\cref{tab:scenes}\else Table~1 of the main paper\fi{} therefore average over smaller homes than the other rows; its time per home includes the failed attempts.

\paragraph{MolmoSpaces.}
We use the homes of its validation split: scenes with a bedroom and a kitchen or living room, and no non-residential room.
\setcounter{table}{\value{apmsavedtable}}
\subsection{Hard constraints}
\label{sec:supp_constraints}
\Cref{tab:supp_constraints} lists the 24 hard constraints.
Each check returns pass, fail or unknown for every element it judges; an element whose geometry, annotation or material is missing is unknown, not passed.
\emph{Hard constr.} in \ifdefined\includeapmsupplement\cref{tab:scenes}\else Table~1 of the main paper\fi{} leaves unknown outcomes out: for each home it is the share of checks with a pass or fail outcome that pass.
Baseline scenes are converted to our home format; for every method, including ours, the floor tolerance of \texttt{furniture-inside-floor} is raised to 0.15\,m, the wall thickness of the converted scenes ($^\dagger$).

\Cref{tab:supp_constraint_fails} gives the share of homes failing each check, among the homes that could be converted to our home format; homes that could not be converted are not included there.
In \emph{Hard constr.} of \ifdefined\includeapmsupplement\cref{tab:scenes}\else Table~1 of the main paper\fi{}, each such home counts as a failure (0\%).
Such homes are 68 of Holodeck's 826 homes with Qwen3.8-27B and 129 of its 778 with Gemma-4-31B: all 68 and 128 of the 129 have overlapping room polygons, which the format does not allow, and one has neither a front door nor an outdoor room.
For MolmoSpaces, 23 of 636 homes are left out: 13 with overlapping rooms, 9 whose static objects end up more than 0.10\,m from their positions in the house file after import, and 1 with an invalid opening.
\begin{table*}[!htbp]
  \centering
  \small
  \setlength{\tabcolsep}{3pt}
  \caption{\textbf{The 24 checks used to assess furnished homes.}
Each row identifies a check, the requirement it enforces, and its tolerance where applicable.
The dagger marks the furniture-placement check whose floor tolerance is 0.15\,m for this evaluation.}
  \label{tab:supp_constraints}
  \begin{tabular}{p{0.27\textwidth}p{0.42\textwidth}p{0.27\textwidth}}
    \toprule
    Check & Requirement & Key thresholds \\
    \midrule
    \multicolumn{3}{l}{\emph{Access}} \\
    \raggedright \texttt{all-\allowbreak rooms-\allowbreak reachable} & \raggedright every room is reachable from the entrance through doors or passages & \raggedright exactly one entrance, leading outside \tabularnewline
    \raggedright \texttt{room-\allowbreak has-\allowbreak entrance} & \raggedright every room has a door or passage & \raggedright a window is no entrance \tabularnewline
    \raggedright \texttt{doors-\allowbreak not-\allowbreak blocked} & \raggedright furniture does not block the walking path through a door or passage & \raggedright walker 0.6\,m (wet), 0.7\,m (interior), 0.8\,m (entrance) \tabularnewline
    \raggedright \texttt{openings-\allowbreak clear} & \raggedright openings stay accessible from the room & \raggedright same walkers \tabularnewline
    \midrule
    \multicolumn{3}{l}{\emph{Fixtures and furniture}} \\
    \raggedright \texttt{bathroom-\allowbreak has-\allowbreak toilet-\allowbreak or-\allowbreak sink} & \raggedright a WC has a toilet and a basin; a full bathroom has a basin and a shower or bath & \raggedright bathroom $\le$4\,m$^2$ counts as WC \tabularnewline
    \raggedright \texttt{home-\allowbreak has-\allowbreak toilet} & \raggedright the home has at least one toilet & \raggedright -- \tabularnewline
    \raggedright \texttt{kitchen-\allowbreak has-\allowbreak sink} & \raggedright every kitchen has a sink & \raggedright -- \tabularnewline
    \raggedright \texttt{bedroom-\allowbreak has-\allowbreak bed} & \raggedright every bedroom has a bed & \raggedright -- \tabularnewline
    \raggedright \texttt{required-\allowbreak room-\allowbreak furniture} & \raggedright every room has the furniture its type needs & \raggedright dining $\ge$2 chairs; small circulation may be bare \tabularnewline
    \midrule
    \multicolumn{3}{l}{\emph{Placement and storage}} \\
    \raggedright \texttt{furniture-\allowbreak inside-\allowbreak floor} & \raggedright floor contact of floor-standing furniture lies on the floor & \raggedright 3\,cm$^\dagger$ \tabularnewline
    \raggedright \texttt{furniture-\allowbreak no-\allowbreak overlap} & \raggedright collision geometry of independent objects does not intersect & \raggedright $\le 10^{-5}$; thin rugs exempt \tabularnewline
    \raggedright \texttt{wall-\allowbreak anchors-\allowbreak touch} & \raggedright wall-anchored furniture touches a wall of its own room & \raggedright gap $\le$1.5\,cm \tabularnewline
    \raggedright \texttt{hanging-\allowbreak attachments-\allowbreak supported} & \raggedright a hanging object's mount contacts a wall or ceiling of its room & \raggedright measured mount point \tabularnewline
    \raggedright \texttt{decoration-\allowbreak supported} & \raggedright decoration rests on the measured surface of its carrier & \raggedright -- \tabularnewline
    \raggedright \texttt{bathroom-\allowbreak sink-\allowbreak rim-\allowbreak supported} & \raggedright a wet-room sink is supported as its model's profile says & \raggedright rim 0.75--0.95\,m \tabularnewline
    \raggedright \texttt{bed-\allowbreak headboard-\allowbreak against-\allowbreak wall} & \raggedright the headboard stands against solid wall, square to it & \raggedright gap $\le$0.10\,m, $\le$2$^\circ$ \tabularnewline
    \raggedright \texttt{seats-\allowbreak face-\allowbreak into-\allowbreak room} & \raggedright a seat faces into the room or its table, not a wall & \raggedright 1\,m free ahead \tabularnewline
    \raggedright \texttt{window-\allowbreak coverings-\allowbreak fit} & \raggedright curtains span their window and hang on its wall & \raggedright 5--30\,cm a side, centred $\le$10\,cm \tabularnewline
    \raggedright \texttt{storage-\allowbreak front-\allowbreak access} & \raggedright storage with doors, drawers or open shelves has free floor in front & \raggedright 0.6\,m \tabularnewline
    \raggedright \texttt{closet-\allowbreak layout-\allowbreak shape} & \raggedright closet storage is one run, an L or a U, against the walls & \raggedright aisle 0.6\,m from the door \tabularnewline
    \raggedright \texttt{closet-\allowbreak storage-\allowbreak occupancy} & \raggedright storage covers enough of a closet's floor & \raggedright $\ge$30\% \tabularnewline
    \midrule
    \multicolumn{3}{l}{\emph{Room program}} \\
    \raggedright \texttt{room-\allowbreak program-\allowbreak plausibility} & \raggedright room area and aspect ratio are plausible for the room type & \raggedright area p1--p99, aspect $\le$p99 of real rooms \tabularnewline
    \raggedright \texttt{no-\allowbreak adjacent-\allowbreak bathrooms} & \raggedright wet rooms are not joined directly by a door or passage & \raggedright a shared wall is allowed \tabularnewline
    \raggedright \texttt{wet-\allowbreak room-\allowbreak floor-\allowbreak not-\allowbreak wood} & \raggedright wet-room floors are tile, stone or concrete & \raggedright -- \tabularnewline
    \bottomrule
  \end{tabular}
\end{table*}
\subsection{Room-layout benchmark}
\label{sec:layout-benchmark}
\Cref{tab:layouts} evaluates the room-layout stage alone, in four blocks.

\paragraph{(a) RPLAN.}
We take 1{,}000 apartments of the Graph2Plan test split of RPLAN.
Every method receives the input it was designed for (bubble diagram or boundary), taken from the same real room layout; our method receives the bubble diagram as its program, bypassing the program stage, once without and once with the real rooms' areas (bubble, bubble+areas), and generates one layout per input.
The bubble diagram holds the rooms and their door connections; the doors are RPLAN's real interior doors, taken from the original RPLAN images because the Graph2Plan version has none.
Where RPLAN draws no door to a room (mostly balconies, 8\% of the apartments), the room gets a door to the room it shares the longest wall with.
ProcTHOR and Infinigen need target room areas and receive the real rooms' areas (bubble+areas).
Holodeck takes text: it receives the same rooms, areas and door connections written as its text query, and we replace its proprietary LLM (GPT-4) by two open-weight models, Qwen3.8-27B and Gemma-4-31B, served locally in bf16 with Holodeck's prompts and settings unchanged; it runs up to three attempts per apartment when its own validity check rejects a floor plan.
Our method and the procedural baselines were not developed on RPLAN.
The learned baselines House-GAN++, HouseDiffusion and WallPlan were trained on RPLAN with their own data splits, which may include some of these 1{,}000 apartments, so their scores may be optimistic.
$N$ is below 1{,}000 where a method produced no layout for an input.

\paragraph{(b--d) Briefs.}
Blocks~(b--d) are the 1{,}000 single-story briefs of \ifdefined\includeapmsupplement\cref{tab:scenes}\else Table~1 of the main paper\fi, split by the layout style our generator drew for each brief (\RSN{} Swiss, \RFN{} Finnish and \RUN{} US homes); each style is compared with real layouts of the matching dataset: Swiss Dwellings~\cite{standfest2022swissdwellings}, CubiCasa5k~\cite{kalervo2019cubicasa5k} and ZInD~\cite{Cruz_2021_CVPR}.
Methods without brief input receive our program (room types, target net areas, room connections).
Holodeck receives the brief itself as its text query (room counts, total net area), with the same language models and attempts as in~(a); as it never sees the program, Graph does not apply to it.
$N$ counts the generated homes; for Holodeck with Qwen3.8-27B it includes two floor plans whose furnishing failed, so its blocks~(b--d) sum to 828 rather than the 826 homes of \ifdefined\includeapmsupplement\cref{tab:scenes}\else Table~1 of the main paper\fi.

\paragraph{Real rows and reference scores.}
The \emph{Real} row of each block scores real room layouts at the same sample size with the same metrics: the real layouts of the 1{,}000 inputs in~(a), and layouts held out from the reference data in~(b--d).
Their FID and KID are the best achievable at that sample size, so we subtract them from every method's score in the same block ($\Delta$FID, $\Delta$KID).
The references are composition-matched real layouts: \RSNRef{} Swiss Dwellings, \RFNRef{} CubiCasa5k and \RUNRef{} ZInD layouts.
Valid of a \emph{Real} row is measured on all layouts of its dataset (\RSNRealAll{}, \RFNRealAll{} and \RUNRealAll{}).

\paragraph{Metrics.}
\emph{Valid}: room layouts without overlap, holes, disconnected rooms, or rooms below minimum area/width or above maximum aspect ratio.
\emph{Reach}: room layouts in which every room is reachable from the entrance through doors or openings.
For the real RPLAN layouts, Reach is measured on the original RPLAN images, which contain the interior doors (the Graph2Plan version does not): a room counts as reachable if it can be walked to from the front door through doors and open passages; most failures are balconies without a drawn door.
Reach is not measurable for the real Swiss Dwellings layouts, which draw no door for open connections between rooms (with doors alone, only 52\% of them would count as fully reachable).
\emph{Types}: room types and counts equal the input.
\emph{Graph}: the door graph equals the input graph (graph edit distance~0).
\emph{Area}: mean relative error of the net room areas.
\emph{$\Delta$FID, $\Delta$KID} ($\times10^3$): FID and KID of room layouts rendered with one shared renderer against the block's reference, minus the score of the \emph{Real} row.
\subsection{Scene metrics and cost}
\label{sec:supp_scene_metrics}
\paragraph{Notes on \ifdefined\includeapmsupplement\cref{tab:scenes}\else Table~1\fi.}
Block~(a) holds scenes published by their authors, not generated from our briefs: the 31 house-level scenes of SceneSmith and 636 homes of the MolmoSpaces validation split (generated with an extended Holodeck); they have no brief, so brief adherence is not defined (--).
Block~(b) holds 1{,}000 single-story homes; every method receives the same briefs, and ProcTHOR and Infinigen Indoors also our room programs.
Holodeck rows cover the briefs for which it produced a valid floor plan within three attempts and a furnished scene (826 with Qwen3.8-27B, 778 with Gemma-4-31B).
\emph{Valid} requires that rooms overlap by at most 2\% of the layout's area, that enclosed gaps belonging to no room cover at most 2\%, that the rooms form one connected component, and that every room meets the minimum area, minimum clear width (largest inscribed circle) and maximum aspect ratio of its type (\cref{tab:supp_valid}).
These are loose fit thresholds (a person or a bed must fit), not a building code, and they are the same for every method and dataset.
\emph{ACC} and \emph{NAV} are computed per room on the oriented object boxes for an agent of radius 0.2\,m: NAV is the share of the room's free floor connected to its doors, ACC the share of floor objects the agent reaches within 0.3\,m.
\emph{COL} counts an object when its footprint overlaps another object's by more than 10\,cm$^2$ and their height ranges overlap by more than 1\,cm; \emph{OOB} counts an object when more than 5\% of its footprint lies outside the rooms.
\emph{Cost}: \$ is the paid API cost per home at sampling time (0 for local code and locally served open-weight models).
Minutes are the wall-clock sampling time of one process per delivered home: the time of every attempt of every brief, failed attempts and briefs without a home included, divided by the delivered homes (\cref{tab:supp_cost}).
Setup, evaluation, rendering and the one-time cost of learning the G-program are not included.
SceneSmith's cost and time are as reported by its authors, per room-level scene; no figure is reported for its house-level scenes.
\begin{table}[!htbp]
  \centering
  \small
  \setlength{\tabcolsep}{4pt}
  \caption{\textbf{Minimum dimensions for a valid room.}
The evaluation applies the same size and shape thresholds to every method.
A dash indicates that the corresponding dimension is unconstrained for that room type.}
  \label{tab:supp_valid}
  \resizebox{\linewidth}{!}{%
  \begin{tabular}{l*{15}{c}}
    \toprule
    & living & kitchen & bedroom & bathroom & dining & study & entrance & corridor & storage & closet & laundry & garage & balcony & outdoor & other \\
    \midrule
    Min.\ area (m$^2$) & 8 & 3 & 5 & 1.5 & 4 & 4 & 1 & 1 & 0.5 & 0.5 & 1 & 10 & 1 & 1 & 0.5 \\
    Min.\ width (m) & 2.4 & 1.5 & 2 & 1 & 2 & 1.8 & 0.9 & 0.8 & 0.6 & 0.6 & 0.9 & 2.4 & 0.8 & 0.8 & 0.6 \\
    Max.\ aspect & 4 & 4 & 3 & 3.5 & 3.5 & 3 & 5 & -- & 6 & 6 & 5 & 4 & 10 & -- & -- \\
    \bottomrule
  \end{tabular}}
\end{table}
\begin{table}[!htbp]
  \centering
  \scriptsize
  \setlength{\tabcolsep}{3pt}
  \caption{\textbf{Time to sample a furnished home,} with the hardware of each sampling process.}
  \label{tab:supp_cost}
  \begin{tabular}{lrrp{0.33\columnwidth}}
    \toprule
    Method & Homes/briefs & min & Hardware per process \\
    \midrule
    ProcTHOR & 1{,}000 / 1{,}000 & 0.33 & AI2-THOR CloudRendering; 4 processes share one NVIDIA A100 \\
    Infinigen Indoors & 1{,}000 / 1{,}000 & 178.4 & one CPU core \\
    Holodeck (Qwen3.8-27B) & 826 / 1{,}000 & 17.2 & local vLLM servers (layout: one H100, 16 processes; furnishing: A100 server, 10 processes); small objects: AI2-THOR on one RTX PRO 5000 \\
    Holodeck (Gemma-4-31B) & 778 / 1{,}000 & 22.4 & local vLLM servers, one per H100 (layout 16, furnishing 8 processes per server); small objects: AI2-THOR on the same H100s \\
    \apm & 1{,}000 / 1{,}000 & 0.39 & one CPU core (x86 servers with 14--60 cores) \\
    \bottomrule
  \end{tabular}
\end{table}

\section{Implementation and reproducibility}
\label{sec:supp_implementation}
\subsection{Program execution and reproducibility}
\label{sec:supp_method_reproducibility}
The main paper describes how agents develop reusable G-programs.
This supplement explains the choices needed to reproduce their outputs, interpret feedback, and compare revisions consistently.
The implementation executes and revises programs, records their outputs and decisions, and rescores stored versions as the objective develops.

\paragraph{Revisiting the same scene.} 
An agent investigating a failure needs to reproduce the choices that produced it.
A seed identifies those choices only when the program and its execution context are unchanged.
We therefore retain the source files, probability tables, catalogue, and engine settings with each G-program version.
The program may span several files, and its probability tables are kept separately so the agents can change how often configurations occur without rewriting the construction logic.
Repeating a seed with this fixed context reproduces the sampled choices.

\paragraph{Scene representation.}
Each scene description contains identifiers linking rooms, architectural elements, and objects.
\Cref{fig:method_stages} groups its contents into the architectural program, room layout, and furnished scene.
The architectural program in (a) describes rooms, their types, target areas, and required connections.
The room layout in (b) adds room geometry, walls, and openings.
Furniture and decoration are shown together in (c), including smaller objects placed on supporting furniture.
Each element also carries provenance, which links it to the G-program commands that created or modified it, as illustrated in (d).

\begin{figure*}[!htbp]
  \centering
  \includegraphics[width=\textwidth]{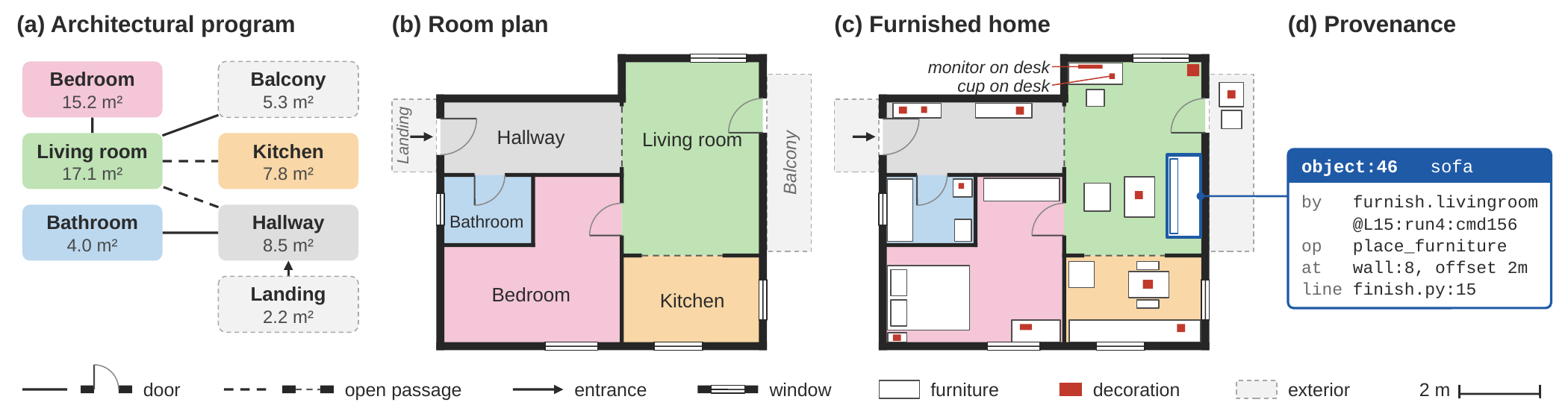}
  \caption{\textbf{Provenance links a furnished scene to the commands that built it.}
    The same scene is shown as (a) required rooms and connections, (b) room geometry with walls and openings, and (c) furniture and decoration.
    Room colors are consistent across panels.
    The illustrative provenance record in (d) links a sofa to its placement command and source location.}
  \label{fig:method_stages}
\end{figure*}

\paragraph{Making construction information available in order.}
The runtime interprets a restricted Python-like language with functions, conditionals, bounded loops, and explicit construction operations.
It executes the architectural program, room layout, furniture, and decoration stages in order. 
This makes room geometry available for furnishing and support surfaces available for decoration.
Programs can schedule work for a later stage; several room-specific programs can contribute within a stage.
Explicit operations allow the runtime to record changes without requiring each generated program to implement its own logging.

\paragraph{Keeping assets at meaningful sizes.}
The catalogue limits which models a G-program can select.
It admits models whose measured extents fall within size bands for their categories, and programs place these models at catalogue size.
Furniture therefore fits a room by selecting a suitable model, not by shrinking one; the language has no scaling operation.

\paragraph{Recording failures as well as completed scenes.} 
The database retains failed attempts, including program errors and timeouts, with the available diagnostic information.
Generation yield is the number of completed scenes divided by the number of attempted seeds.
Keeping that denominator matters: a revision that simply fails on difficult seeds has not solved those cases.
Execution time is recorded separately to assess the cost of producing scenes, using controlled benchmarks for candidate comparisons.

\subsection{Turning requirements into useful feedback}
\label{sec:supp_method_feedback}
The evaluator must distinguish an unwanted generated pattern from ordinary variation in the reference population.
It must also accommodate requirements that the user wants even when real homes do not always satisfy them.
We therefore retain the origin of each comparison: a reference statistic, a stated design target, or a geometric constraint.

\paragraph{Incomplete reference data.}
Reference sources are chosen by property and housing population.
For example, layout observations and furniture observations need not come from the same homes.
An unannotated category cannot support a presence-rate comparison.
Measurements use only eligible observations, with sample counts retained alongside the values.
Statistical comparisons generally require at least 100 generated observations; for specified HSSD references, at least 60 reference observations. 
The configuration records these thresholds and reference-noise adjustments so small reference collections are not treated as exact targets.
Versions are compared on measurements active for both, and a comparison that loses a substantial share of its previously available observations receives the maximum discrepancy, so a revision cannot improve its score by making a comparison unavailable.
Newly supported output properties have a separate acceptance criterion rather than counting their earlier absence as a perfect match. 

\paragraph{Choosing a discrepancy.}
Rates and means use relative differences, numerical distributions use Kolmogorov--Smirnov distances, and categorical distributions use total variation distances.
Each comparison is mapped to a discrepancy in $[0,1]$ before aggregation.
Some user requirements instead define acceptable bands, such as a preference for more densely furnished scenes.
Their target values express the request rather than an estimate of how frequently the property occurs in real homes.
The scoring configuration fixes the comparison, normalization, reference subset, and weight.
Check results first supply aggregate frequencies, which receive explicitly defined discrepancies against reference values or design targets.
For example, the circulation discrepancy is the fraction of rooms with disconnected walkable regions divided by 0.10, capped at one.
Yield and the hard-constraint pass rate instead have discrepancies equal to one minus their success fractions.
For aggregation, components are grouped by topic: their weighted agreement scores are normalized within each group over available components, then combined using group weights normalized over contributing groups.
The main paper's $\alpha_c$ incorporates both normalizations; inactive or unavailable terms have zero weight. 

\paragraph{Deciding which checks affect selection.}
A visual finding first motivates a candidate check.
The reference-based admission procedure requires enough real observations (at least 30), a Wilson upper 95\% bound on the real failure rate of at most 0.10, and a higher failure rate on generated data under a one-sided Fisher test with Holm correction at 5\%.
This guards against treating a common real configuration as a defect merely because it looked unusual in a rendered example.
The Orchestrator can also admit an explicit requirement or an asset-specific defect check with recorded justification when a reference frequency is not the relevant criterion.
Such checks are identified as requirements, rather than described as statistically discovered properties.
Checks can remain diagnostic without blocking a revision, and can be retired when invalid or explicitly withdrawn.

\paragraph{Constraint satisfaction.}
The hard constraints of the task specification (24 checks in our experiments, \cref{sec:supp_constraints}) enter the quality score through the hard-constraint pass rate: the fraction of completed scenes for which every hard-constraint check returns pass; an unknown result is not a pass.
Further checks can be kept as diagnostics without affecting selection.
Generation yield and the hard-constraint pass rate both have a target of one, but they answer different questions: whether a scene was produced, and whether the completed scene meets the requirements.
When versions are compared, both are graded with the same checks.

\subsection{Using provenance to guide revisions}
\label{sec:supp_method_provenance}
The source links in the main paper's provenance figure narrow the code an agent needs to inspect.
For each entity, the database records its creation command, subsequent modifying commands, and most recent command.
Command records include the source file and line, helper-call stack, and links to earlier commands that scheduled the work.
Every seeded random draw is also recorded with its arguments and sampled value.
Queries can join check results to affected entities and group the associated commands by source location across scenes. 

This is execution provenance, not complete data-flow tracking.
Reading a probability table or querying existing geometry does not record every dependency of the eventual placement.
The emitting line and its callers therefore identify where to investigate, while replay and source inspection help the agent examine the choices that led there.
The proposed explanation is tested by generating and checking another batch after the edit.

The system supports both aggregate reports and direct provenance queries.
In development, failures were grouped across hundreds of seeds by traced source line; other revisions relied more on statistical comparisons and visual findings.
The mechanism does not require every revision to begin with a provenance query.

\subsection{Orchestration and candidate selection}
\label{sec:supp_method_selection}
The Orchestrator coordinates development in an interactive LLM session, while the Generator and Discriminator Agents carry out assigned tasks with their own contexts and tools.
User messages, agent messages, and completed jobs trigger new decisions.
The Orchestrator can pause or redirect delegated work and ask the user to resolve an unclear preference.
The answer becomes a new event from which development resumes.
There is no fixed generator--discriminator alternation or automatic convergence test; development can stop, wait for input, or resume when further work is requested.

\paragraph{Revising the objective.}
New measurements are proposed by the Orchestrator or the Discriminator Agent.
The Discriminator Agent proposes the numerical weights and implements scoring changes; the Orchestrator approves and versions the scoring function against the user's guidance.
The weights are chosen explicitly, not fitted from data.
Visual inspection supplies evidence for these decisions but does not directly overwrite the numerical score.
To change the score, a finding must be expressed through a measurement or target and incorporated into the versioned scoring configuration.
This distinction keeps a repeatable executable objective while allowing the user to clarify what should matter.

\paragraph{Comparing program revisions.}
A few improved examples are not enough to select a reusable generator. Each proposed revision is stored as a separate candidate, so testing it does not replace the accepted version.
With the quality score, a candidate is compared with its parent on the same seeds and the same D-program configuration: paired bootstrap resampling of the seeds (400 resamples) gives a 95\% interval for the score difference, whose lower bound must exceed zero, and regression checks on validity, individual components, validated defects and execution time are corrected for multiple testing in the same way as the admission tests of \cref{sec:supp_method_feedback}.
A candidate can also be accepted on targets the user sets for individual components of the quality score.
The development run of the main paper (Sec.~4.3) used this criterion: because its hard constraints were extended during the run, candidates were accepted on the hard-constraint pass rate and the generation yield on the development seeds, and agreement with the reference data (Fig.~5) was checked separately.
A rejected candidate leaves the accepted generator unchanged; a later candidate built on its idea is again judged against a named accepted parent, and parallel work is combined into one candidate that is tested before it is accepted.

\paragraph{Other reasons to carry a revision forward.}
A validated defect fix can also qualify when its targeted defect rate improves, the aggregate score is unchanged within uncertainty, and the remaining checks hold.
Requested changes, newly supported capabilities, neutral changes to random streams and explicit Orchestrator decisions have separate recorded criteria and reasons.
Each decision is stored with the scoring version used at the time, so rescoring an old generator under a later objective does not change why it was selected, and the development sequence need not be monotonic under a later score.


\fi
\end{document}